\documentclass[10pt,twocolumn,letterpaper]{article}

\usepackage{cvpr}            
\usepackage[accsupp]{axessibility}  

\usepackage{amssymb}
\usepackage{amsmath}
\usepackage{amsfonts}
\usepackage{tabularx}
\usepackage{booktabs}
\usepackage{tabularx}
\usepackage{multirow}
\usepackage{subcaption}
\usepackage{xcolor}
\usepackage{graphicx}
\usepackage{verbatim}
\usepackage{tikz,pgfplots}
\usepackage{bbm}

\newcommand{\PAR}[1]{\vskip4pt \noindent {\bf #1~}}
\newcommand{\PARbegin}[1]{\noindent {\bf #1~}}
\newcommand{\tablestep}[1]{\textcolor{gray!95}{(#1)}}

\newcommand{\eomt}{EoMT}
\newcommand{\videomt}{VidEoMT}

\usepackage{xspace}

\usepackage{wrapfig}

\newcommand{\downrightarrow}{\hspace{0pt}%
  \raisebox{1.5pt}{\begin{tikzpicture}[scale=0.4, baseline=(current bounding box.south)]
    \draw[-latex] (0.0, 0.5) -- (0.0, 0.25) -- (0.55, 0.25);
  \end{tikzpicture}} %
}

\expandafter\def\expandafter\normalsize\expandafter{%
    \normalsize%
    \setlength\abovedisplayskip{2pt}%
    \setlength\belowdisplayskip{2pt}%
    \setlength\abovedisplayshortskip{2pt}%
    \setlength\belowdisplayshortskip{2pt}%
}

\newcolumntype{Y}{>{\centering\arraybackslash}X}

\definecolor{c_red}{HTML}{ea4335}
\definecolor{c_green}{HTML}{34a853}

\colorlet{tablered}{c_red!99!black}
\colorlet{tablegreen}{c_green!99!black}

\newcommand{\qincnew}[1]{\textcolor{tablegreen}{\tiny \textsuperscript{$\uparrow$#1}}}
\newcommand{\qdecnew}[1]{\textcolor{tablered}{\tiny \textsuperscript{$\downarrow$#1}}}

\definecolor{cvprblue}{rgb}{0.21,0.49,0.74}
\usepackage[pagebackref,breaklinks,colorlinks,allcolors=cvprblue]{hyperref}
\newcommand\blfootnote[1]{%
  \begingroup
  \renewcommand\thefootnote{}\footnote{#1}%
  \addtocounter{footnote}{-1}%
  \endgroup
}

\usepackage{amsmath,amsfonts,bm}

\def\eqref#1{equation~\ref{#1}}

\def\1{\bm{1}}

\DeclareMathAlphabet{\mathsfit}{\encodingdefault}{\sfdefault}{m}{sl}
\SetMathAlphabet{\mathsfit}{bold}{\encodingdefault}{\sfdefault}{bx}{n}

\title{VidEoMT: Your ViT is Secretly Also a Video Segmentation Model}

\author{
Narges Norouzi\textsuperscript{1}\qquad 
Idil Esen Zulfikar\textsuperscript{2,\dag} \qquad 
Niccolò Cavagnero\textsuperscript{1,\dag} \qquad 
Tommie Kerssies\textsuperscript{1} \\
Bastian Leibe\textsuperscript{2} \qquad 
Gijs Dubbelman\textsuperscript{1} \qquad 
Daan de Geus\textsuperscript{1}\\[0.6em]
$^1$Eindhoven University of Technology\qquad $^2$RWTH Aachen University \\
}

\begin{document}
\maketitle
\begin{abstract}
Existing online video segmentation models typically combine a per-frame segmenter with complex specialized tracking modules. While effective, these modules introduce significant architectural complexity and computational overhead. Recent studies suggest that plain Vision Transformer (ViT) encoders, when scaled with sufficient capacity and large-scale pre-training, can conduct accurate image segmentation without requiring specialized modules. Motivated by this observation, we propose the \textit{Video Encoder-only Mask Transformer (VidEoMT)}, a simple encoder-only video segmentation model that eliminates the need for dedicated tracking modules. To enable temporal modeling in an encoder-only ViT, VidEoMT introduces a lightweight query propagation mechanism that carries information across frames by reusing queries from the previous frame. To balance this with adaptability to new content, it employs a query fusion strategy that combines the propagated queries with a set of temporally-agnostic learned queries. As a result, VidEoMT attains the benefits of a tracker without added complexity, achieving competitive accuracy while being 5$\times$--10$\times$ faster, running at up to 160 FPS with a ViT-L backbone.
Code: \href{https://tue-mps.github.io/videomt/}{https://tue-mps.github.io/videomt/}.
\end{abstract}

\blfootnote{{$^\dag$Equal contribution.}}\vspace{-1em}
\section{Introduction}
\label{sec:intro}

The \textit{video segmentation} task involves segmenting and classifying objects in each frame, while also matching those objects across frames. As such, a video segmentation model should have the capability to \textit{localize} objects, \textit{classify} them, and \textit{track} them across different frames. For this reason, great progress has been made through the introduction of specialized neural network components that are designed to improve one or more of these capabilities. Current methods obtain state-of-the-art performance by combining many such specialized components within increasingly complex models~\cite{lee2025cavis,zhang2023dvis,zhang2025dvis++,zhou2024dvisdaq}, building upon years of prior work. This trend of increasing complexity motivates us to explore whether such complexity is necessary, or if this task can be solved with similar accuracy using a simpler approach.
\definecolor{caviscolor}{HTML}{397FBE}
\colorlet{eomtcaviscolor}{caviscolor!50!white}
\definecolor{videomtcolor}{HTML}{ff6d01}
\begin{figure}[t!]
    \begin{tikzpicture}
        \def\vitgmark{square*}
        \def\vitLmark{*}
        \def\vitBmark{triangle*}
        \def\vitSmark{diamond*}
        \def\cavisLdata{(15, 68.9)}
        \def\cavisLdataX{15} %
        \def\cavisLdataY{68.9} %
        \def\cavisBdata{(18, 58.8)}
        \def\cavisSdata{(19, 54.8)}
        \def\eomtcavisLdata{(42, 68.1)}

        \def\eomtcavisBdata{(66.5, 57.4)}
        \def\eomtcavisSdata{(93.4, 50.3)}

        \def\videomtLdataX{160} %
        \def\videomtLdataY{68.6} %
        \def\videomtLdata{(160, 68.6)}
        \def\videomtBdata{(251, 58.2)}
        \def\videomtSdata{(294, 52.8)}
    
        \begin{axis}[
            font=\small,
            width=1.05\linewidth,
            height=0.73\linewidth,
            xmode=log,
            xlabel={Frames per second (FPS) [log scale]},
            ylabel={Average Precision (AP)},
            ylabel style={yshift=-1.25em, xshift=0em},
            xlabel style={yshift=0.6em},
            xmin=10, xmax=360,
            ymin=38, ymax=74,
            xtick={20, 40, 80, 160, 320},
            xticklabels={20, 40, 80, 160, 320},
            ytick={46, 54, 62, 70},
            legend style={font=\scriptsize, at={(0.01,0.01)}, anchor=south west},
            legend cell align=left,
            ymajorgrids=true,
            xmajorgrids=true,
            grid style=dashed,
            tick label style={font=\footnotesize},
            label style={font=\footnotesize},
        ]

    \addlegendimage{empty legend}
    \addlegendentry{\hspace{-12pt}%
    \tikz \node[mark size=2pt, inner sep=0.8pt] at (0, 0) {\pgfuseplotmark{\vitLmark}};~~ViT-L~~
    \tikz \node[mark size=2pt, inner sep=0.8pt] at (0, 0) {\pgfuseplotmark{\vitBmark}};~~ViT-B~~
    \tikz \node[mark size=2pt, inner sep=0.8pt] at (0, 0) {\pgfuseplotmark{\vitSmark}};~~ViT-S};

    \addlegendimage{color=caviscolor,
            mark=*,
            line width=0.75pt}
    \addlegendentry{CAVIS}
    \addlegendimage{color=eomtcaviscolor,
        mark=*,
        line width=0.75pt}
    \addlegendentry{EoMT + CAVIS}
    \addlegendimage{color=videomtcolor,
            mark=*,
            line width=0.75pt}
    \addlegendentry{\textbf{VidEoMT (Ours)}}

    \tikzset{cavis line style/.style={color=caviscolor,line width=0.3mm}}

    \addplot[cavis line style, mark=\vitLmark, mark size=2.5pt] coordinates {\cavisLdata};
    \addplot[cavis line style, mark=\vitBmark, mark size=2.5pt] coordinates {\cavisBdata};
    \addplot[cavis line style, mark=\vitSmark, mark size=2.5pt] coordinates {\cavisSdata};
    \addplot[cavis line style,
            mark=none,
    ]
    coordinates {
            \cavisLdata
            \cavisBdata
            \cavisSdata
    };

    \tikzset{eomtcavis line style/.style={color=eomtcaviscolor,line width=0.3mm}}

    \addplot[eomtcavis line style, mark=\vitLmark, mark size=2.5pt] coordinates {\eomtcavisLdata};
    \addplot[eomtcavis line style, mark=\vitBmark, mark size=2.5pt] coordinates {\eomtcavisBdata};
    \addplot[eomtcavis line style, mark=\vitSmark, mark size=2.5pt] coordinates {\eomtcavisSdata};
    \addplot[eomtcavis line style, densely dashed,
            mark=none,
    ]
    coordinates {
            \eomtcavisLdata
            \eomtcavisBdata
            \eomtcavisSdata
    };

    \tikzset{videomt line style/.style={color=videomtcolor,line width=0.3mm}}

    \addplot[videomt line style, mark=\vitLmark, mark size=2.5pt] coordinates {\videomtLdata};
    \addplot[videomt line style, mark=\vitBmark, mark size=2.5pt] coordinates {\videomtBdata};
    \addplot[videomt line style, mark=\vitSmark, mark size=2.5pt] coordinates {\videomtSdata};
    \addplot[videomt line style,
            mark=none,
    ]
    coordinates {
            \videomtLdata
            \videomtBdata
            \videomtSdata
    };

    \draw[-stealth, bend left=5, dashed, thick] 
      (axis cs:\cavisLdataX+1,\cavisLdataY+0.1)
      to [bend left=5] 
      node[pos=0.56, above, font=\small] {$>10\times$ faster}
      (axis cs:\videomtLdataX-8,\videomtLdataY+0.1);
    
    \end{axis}

\end{tikzpicture}
    \caption{\textbf{CAVIS vs.\ \videomt{} (Ours).} VidEoMT is much faster than both CAVIS~\citep{lee2025cavis} and a combination of EoMT~\cite{kerssies2025eomt} and CAVIS, while maintaining competitive AP across different sizes of DINOv2~\cite{oquab2023dinov2}. Evaluated on  YouTube-VIS 2019 \textit{val}~\cite{yang2019video}.}
    \label{fig:speed_plot_teaser}
\end{figure}

\begin{figure*}[ht!]
\centering
\includegraphics[width=1.0\linewidth]{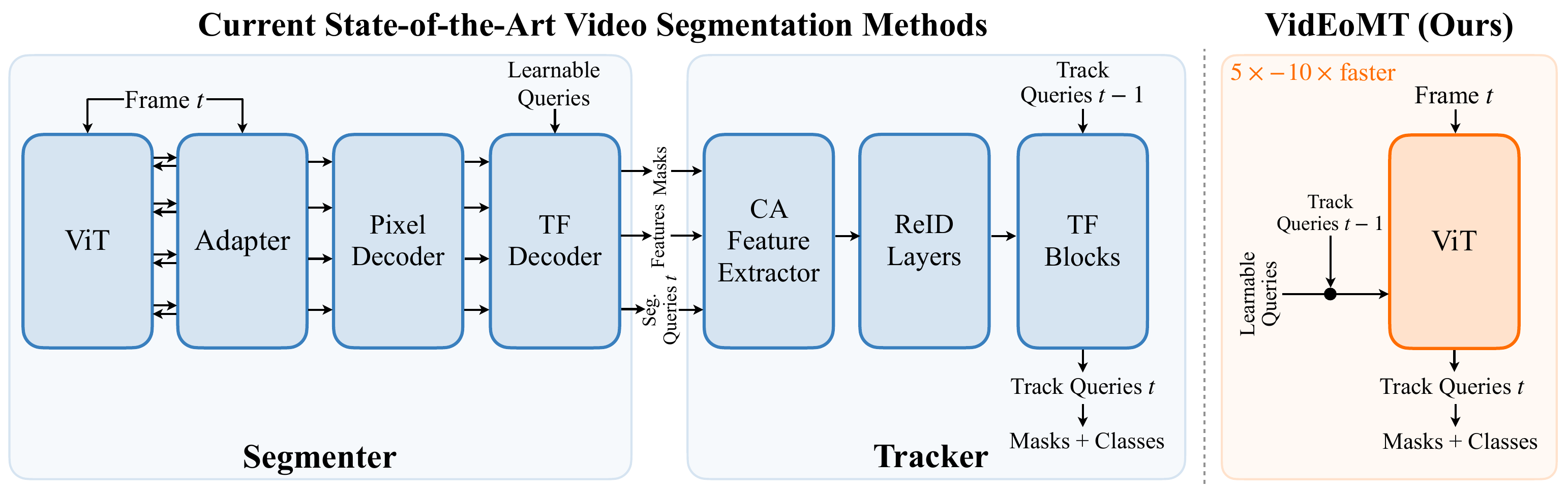}
\caption{\textbf{Current State-of-the-Art Video Segmentation Methods vs.\ \videomt{} (Ours).} We compare the architectures of current state-of-the-art video segmentation methods -- using CAVIS~\cite{lee2025cavis} as a representative example -- and our encoder-only VidEoMT method. VidEoMT streamlines the video segmentation framework, relying on the power of large-scale pre-training with vision foundation models rather than handcrafted task-specific components. \textit{TF} means \textit{Transformer} and \textit{CA} means \textit{context-aware}.}
\label{fig:teaser}
\end{figure*}

In this paper, we hypothesize that a simpler approach can match the accuracy of more complex models by making use of powerful vision foundation models (VFMs). These VFMs~\cite{simeoni2025dinov3, oquab2023dinov2, fang2024eva02}, which typically adopt the Vision Transformer (ViT) architecture~\cite{dosovitskiy2021vit}, are pre-trained on large amounts of data, and have proven to be solid foundations for subsequent finetuning for downstream tasks. For video segmentation, VFMs like DINOv2~\cite{oquab2023dinov2} are incorporated in most recent models~\cite{zhang2025dvis++, lee2025cavis, zhou2024dvisdaq} by being extended with many specialized components. However, we believe that these strong pre-trained ViTs can learn to take over many of the functionalities of the specialized components that are typically added on top, making these components \emph{redundant}. This is inspired by Kerssies \etal~\cite{kerssies2025eomt}, who showed that state-of-the-art image segmentation can be achieved by simply adding a few learnable queries to a large pre-trained ViT with a model called \eomt, without needing specialized components. EoMT demonstrates that a large, pre-trained ViT encoder can learn to effectively localize and classify objects, which is also required for video segmentation. However, video segmentation has an additional requirement: temporally tracking objects across frames.

We expect that the pre-trained ViT encoders from VFMs will also be able to learn to track objects because of the training objectives of these VFMs. 
For instance, DINO-style models~\cite {caron2021dino,oquab2023dinov2,simeoni2025dinov3} employ training objectives that promote consistent feature representations for a given object across different views.
Cross-view consistent features are crucial for tracking, as they allow for the identification of the same object in different frames.
This makes pre-trained ViTs from VFMs highly suited for video segmentation.

We verify our hypothesis about the redundancy of complex components in video segmentation models by taking state-of-the-art video segmentation models~\cite{zhang2025dvis++, lee2025cavis, zhou2024dvisdaq} and evaluating the effect of removing their specialized modules. These existing models all follow roughly the same paradigm: they first employ a \textit{segmenter}, which predicts frame-level segmentation masks and class labels and outputs object-level feature queries, and then apply a \textit{tracker} to match object-level feature queries across different video frames. As shown in \cref{fig:teaser} (left) for the example of CAVIS \cite{lee2025cavis}, both the segmenter and the tracker consist of many specialized components. We first replace the complex segmenter with \eomt~\cite{hendrycks2016gelu}, followed by a step-by-step removal of specialized tracking modules.

Next, we move away from the conventional decoupling of segmenter and tracker, and instead explore whether temporal modeling can be conducted within a ViT encoder. To this end, we introduce a lightweight approach based on (1) \textit{query propagation}, where object-level queries are carried across frames to enable temporal modeling in an encoder-only framework, and (2) \textit{query fusion}, which combines propagated queries with learned queries to allow the identification of newly appearing objects. This leads to the design of the \textit{Video Encoder-only Mask Transformer} (VidEoMT), which unifies segmentation and temporal association within the ViT, as illustrated in \cref{fig:teaser} (right).

By no longer requiring complex specialized components and performing all computations within a single ViT-style model, VidEoMT is remarkably efficient. Through experiments, we find that VidEoMT with a ViT-Large backbone is over $10\times$ faster than existing state-of-the-art methods on the YouTube-VIS~\cite{yang2019video} benchmarks, achieving processing speeds of up to 160 FPS, as shown in \cref{fig:speed_plot_teaser}. Importantly, this speed is obtained while maintaining a comparable accuracy. These findings are further validated on the VIPSeg~\cite{miao2022large} and VSPW~\cite{miao2021vspw} benchmarks, where VidEoMT consistently achieves speedups of 5$\times$--10$\times$ with negligible impact on accuracy.
Such speed-up factors can be a veritable game changer for applications, enabling online video processing across a wide range of use cases. 
These results also validate our hypothesis that a large, extensively pre-trained ViT can take over the functionalities of specialized components to conduct accurate video segmentation, without requiring additional complex components.

In summary, we make the following contributions:

 \begin{itemize}
 \item We propose VidEoMT, a simple and highly efficient architecture for video segmentation, that unifies segmentation and temporal association within a single ViT encoder. 
 \item Using VidEoMT, we demonstrate that a sufficiently large, pre-trained ViT can learn to take over the functionality of specialized components for video segmentation. 
 \item We show that VidEoMT, with its simple encoder-only architecture, can achieve accuracies comparable to the state of the art while being up to $10\times$ faster.  
\end{itemize}

\section{Related Work}
\label{sec:related}

\PARbegin{Image Segmentation.}
Image segmentation requires that objects in an image are segmented and classified. 
Early image segmentation models treated this task as a \textit{per-pixel classification} problem, predicting a class label for each pixel~\cite{chen2018deeplab,chen2018deeplabv3plus,long2015fcn}. 
Later works propose an alternative \textit{mask classification} approach, where a model predicts a \textit{segment} -- consisting of a segmentation mask and class label -- for each object in the image~\cite{cheng2021per}.
These mask classification methods typically make use of Mask Transformers, which use image features from a backbone and learnable queries to predict a segmentation mask and class label for each query with a Transformer decoder~\cite{wang2021maxdeeplab,cheng2022mask2former,jain2023oneformer,cavagnero2024pem}.
Recently, EoMT~\cite{kerssies2025eomt} has demonstrated that it is possible to conduct accurate image segmentation without a decoder or other task-specific components, by simply feeding the learnable queries directly into a large, pre-trained ViT.
In this work, inspired by EoMT, we investigate whether video segmentation models can be simplified in a similar manner, with the goal of improving efficiency while preserving high accuracy.

\PAR{Video Segmentation.} Video segmentation is a well-established computer vision task, encompassing video instance segmentation (VIS)~\cite{yang2019video}, video panoptic segmentation (VPS)~\cite{kim2020video}, and video semantic segmentation (VSS)~\cite{nilsson2018semantic}, where the primary objective is to segment, classify, and track all objects of interest in a video. %
Current VIS, VPS, and VSS methods typically use Mask Transformer-based architectures~\cite{heo2022vita, huang2022minvis, heo2023generalized, zhang2023dvis, zhang2025dvis++, lee2025cavis, zhou2024dvisdaq, weng2023mask, shin2024video, cheng2021mask2formervis}. They extend Mask Transformers for image segmentation~\cite{cheng2022mask2former} into the video domain by incorporating specialized tracking components or enhancing temporal representations. 
The most recent methods~\cite{huang2022minvis, zhang2023dvis, zhang2025dvis++, lee2025cavis, zhou2024dvisdaq} are \emph{universal} models, which can handle VIS, VPS, and VSS within a single framework.
These models follow a decoupled paradigm, where the segmentation and tracking sub-tasks are separated.
First, a segmenter conducts image segmentation for each frame, and then a tracker associates these segmented objects over time.
Generally, both the segmenter and the tracker contain various specialized components, which increase accuracy but reduce efficiency.
In this work, we analyze these universal video segmentation models and demonstrate that they can be simplified to an encoder-only design, significantly improving efficiency while achieving competitive accuracy.

\section{Method}

\label{sec:method}

\subsection{Task Definition}

We consider the task of \emph{online video segmentation}, where the goal is to assign a class label and binary mask to each object in every frame, while also associating predictions of the same object across time steps to ensure temporal consistency. Here, we use the term ``object" broadly to refer to either object instances (as in VIS), semantic classes (as in VSS), or both (as in VPS).

Formally, a video is a sequence of $T$ frames $\mathcal{V} = \{\mathbf{I}_1, \mathbf{I}_2, \dots, \mathbf{I}_T\}$. For each frame $\mathbf{I}_t \in \mathbb{R}^{3 \times H \times W}$ with spatial resolution $(H, W)$, a model should yield a set of $K_t$ predictions $\mathcal{Y}_t = \{(\mathbf{m}_{t,i}, c_{t,i}) \}_{i=1}^{K_t}$,
where $\mathbf{m}_{t,i} \in \{0,1\}^{H \times W}$ is a binary segmentation mask, and $c_{t,i} \in \{1, \dots, C\}$ is a semantic category label from $C$ classes. Additionally, these per-frame predictions must be temporally associated across frames to maintain identity consistency. That is, each prediction $(\mathbf{m}_{t,i}, c_{t,i})$ at time $t$ should be matched to a corresponding prediction in a previous frame at $t-1$ if they refer to the same object.
The task must be solved in an \emph{online} manner: at timestep $t$, predictions $\mathcal{Y}_t$ may only depend on the current frame $\mathbf{I}_t$ and earlier frames $\{\mathbf{I}_1, \dots, \mathbf{I}_{t-1}\}$.

\subsection{Preliminaries}

Current state-of-the-art online video segmentation models~\cite{zhang2023dvis,zhang2025dvis++,lee2025cavis,zhou2024dvisdaq} typically decompose the video segmentation pipeline into two distinct components: a \textit{segmenter}, which is responsible for generating segmentation masks and class labels for each frame, and a \textit{tracker} that ensures temporal association by linking the segmenter's predictions across frames, associating object instances over time (see \cref{fig:teaser}).

\PAR{Segmenter.}
The segmenter $\mathcal{S}$ generates frame-level image segmentation predictions, yielding a class label and a binary mask for each object. To obtain these predictions, state-of-the-art methods combine a pre-trained ViT~\cite{dosovitskiy2021vit,oquab2023dinov2}, a ViT-Adapter~\cite{chen2023vitadapter}, and a Mask Transformer segmentation decoder~\cite{cheng2022mask2former}. 
The ViT encoder embeds an input image $\mathbf{I}_{t}$ into non-overlapping patch tokens and processes them with $L$ Transformer blocks. A CNN-based ViT-Adapter augments the encoder with multi-scale features, which are fused and refined in the Mask2Former~\cite{cheng2022mask2former} head by a pixel decoder, producing a set of enriched features $\{\mathbf{F}_{4}, \mathbf{F}_{8}, \mathbf{F}_{16}, \mathbf{F}_{32}\}$, with $\mathbf{F}_{i} \in \mathbb{R}^{D \times (H/i) \times (W/i)}$. A Transformer decoder then updates $N$ learnable queries $\mathbf{Q}^\textrm{lrn} = \{ \mathbf{q}^\textrm{lrn}_i \in \mathbb{R}^{D}\}_{i=1}^{N}$ through cross- and self-attention, yielding refined queries $\mathbf{Q}^\mathcal{S} = \{\mathbf{q}^\mathcal{S}_i \in \mathbb{R}^{D}\}_{i=1}^{N}$. These refined queries, in combination with the feature maps $\mathbf{F}_4$, are used to generate the final segmentation outputs: class labels are predicted via a linear layer applied to each query, while binary masks are produced by passing the queries through a three-layer MLP followed by a dot product with the pixel-level feature maps.

\PAR{Tracker.}The objective of the tracker $\mathcal{T}$ is to associate the segmenter's predictions across frames to maintain consistent object identities over time. Rather than relying on the predicted masks and class labels from the segmenter, the tracker performs association based on the per-object query embeddings $\mathbf{Q}^\mathcal{S}$. Formally, the tracker $\mathcal{T}$ aligns the query embeddings from the current frame, $\mathbf{Q}^{\mathcal{S}}_{t}$, with the temporally updated queries from the previous frame, $\mathbf{Q}^{\mathcal{T}}_{t-1}$, to achieve correspondence between object instances over time, producing temporally updated queries $\mathbf{Q}^{\mathcal{T}}_{t}$:
\begin{equation}
    \mathbf{Q}^{\mathcal{T}}_{t} = \mathcal{T}\!\left(\mathbf{Q}^{\mathcal{S}}_{t}, \mathbf{Q}^{\mathcal{T}}_{t-1}\right).
\end{equation}
In practice, $\mathcal{T}$ consists of $L$ Transformer blocks with cross-attention, self-attention, and feed-forward layers. During cross-attention, queries $\mathbf{Q}^{\mathcal{T}}_{t-1}$ serve as \textit{queries} and $\mathbf{Q}^{\mathcal{S}}_{t}$ as \textit{keys} and \textit{values}, allowing the tracker to align and update the representations of identical objects across consecutive frames. The output of the cross-attention layer is then refined by a self-attention layer, which further enhances temporal coherence among the updated queries. These operations ensure that a consistent query ordering is obtained, meaning that $\mathbf{Q}^{\mathcal{T}}_{t}$ represents the same objects as $\mathbf{Q}^{\mathcal{T}}_{t-1}$, in the same order. The refined queries $\mathbf{Q}^{\mathcal{T}}_{t}$ can then be used to predict temporally consistent masks and class labels, following the same procedure as in the segmenter.

\PAR{Context-Aware Features.}
To enrich query embeddings $\mathbf{Q}^\mathcal{S}_t$ with information from the local neighborhood of each object, the state-of-the-art method CAVIS~\cite{lee2025cavis} introduces \emph{context-aware features}. Concretely, given predicted masks $\mathbf{M}_t=\{\mathbf{m}_{t,i}\}_{i=1}^K$ and features $\mathbf{F}_{4,t}$ at timestep $t$, binary boundary maps $\mathbf{B}_{t,i} \in \{0,1\}^{(H/4) \times (W/4)}$ are extracted using a Laplacian filter. %
Next, the features $\mathbf{F}_{4,t}$ are smoothed with an average filter, yielding $\mathbf{F}^\mathcal{A}_{4,t}$.
Finally, the context-aware features $\mathbf{Q}^{\mathcal{A}}_t = \{\mathbf{q}^{\mathcal{A}}_{t,i} \in \mathbb{R}^{D}\}_{i=1}^N$, extracted by pooling the smoothed features at boundary pixels, are concatenated with the per-frame query embeddings $\mathbf{Q}^\mathcal{S}_{t}$. This process produces an enriched set of queries $\mathbf{Q}^\mathcal{C}_{t} = \{\mathbf{q}^{\mathcal{C}}_{t,i} \in \mathbb{R}^{2D}\}_{i=1}^N$, which are then fed into the tracker's Transformer blocks in place of the segmenter queries $\mathbf{Q}^{\mathcal{S}}_{t}$.

\PAR{Re-identification Layers.} To further improve robustness, recent methods~\cite{zhang2025dvis++,zhou2024dvisdaq,lee2025cavis} employ \emph{re-identification layers}. 
These layers are paired with contrastive training objectives to enforce similarity between embeddings of the same instance while separating those of different instances. In practice, query embeddings $\mathbf{Q}^\mathcal{S}_{t}$ from the segmenter are usually fed to the re-identification layers, implemented as a 3-layer $\texttt{MLP}$.
In CAVIS, this MLP is instead applied to the context-aware queries $\mathbf{Q}^{\mathcal{C}}_{t}$:
\begin{equation}
\mathbf{Q}^\mathcal{R}_{t} = 
\texttt{MLP}\!\left(\mathbf{Q}^\mathcal{C}_{t}\right).
\end{equation}
This yields enhanced queries $\mathbf{Q}^\mathcal{R}_{t}$ which are subjected to contrastive learning and are fed into the tracker's Transformer blocks, in place of the context-aware queries $\mathbf{Q}^\mathcal{C}_{t}$.

\begin{figure*}[t]
\centering
\includegraphics[width=1.0\linewidth]{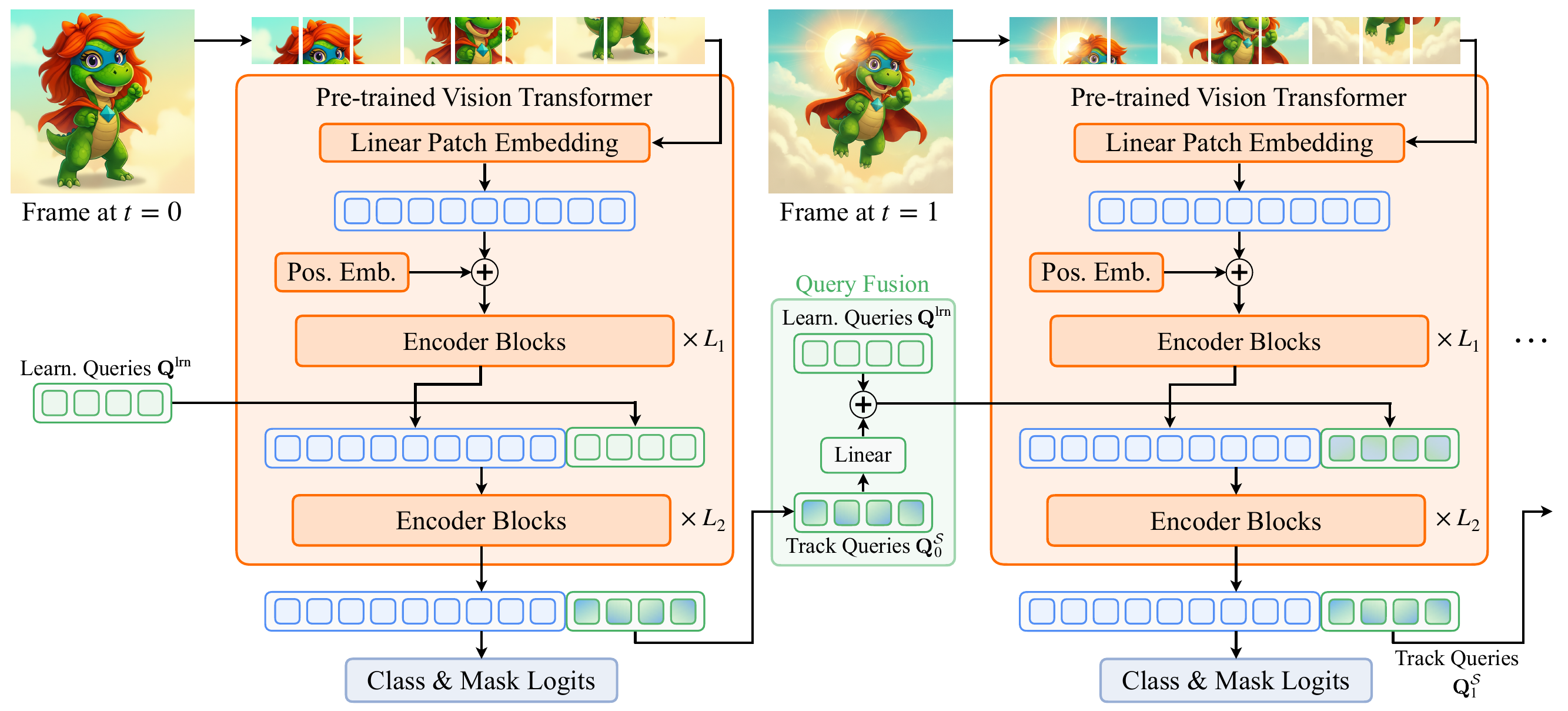}
\caption{\textbf{\videomt{} architecture.} For the initial video frame at $t = 0$, learnable queries are concatenated to the patch tokens after the first $L_1$ ViT blocks. Both sets of tokens are then jointly processed in the last $L_2$ blocks, outputting predictions and track queries. For successive frames, learnable queries and previous-frame track queries are fed to the \textit{query fusion} module before being processed by the ViT blocks.}
\label{fig:arch}
\end{figure*}

\subsection{Removing Task-specific Components}
\label{sec:remove_task_sepc_comp}

Recently, EoMT~\cite{kerssies2025eomt} has challenged the dominant paradigm in image segmentation that relies on many specialized components, showing that this task can be performed in an encoder-only fashion, given a sufficiently large ViT model and strong pre-training. This is also relevant in the video domain, as the segmenter modules of state-of-the-art video segmentation models also use the same specialized components. In EoMT~\cite{kerssies2025eomt},  learned queries $\mathbf{Q}^\textrm{lrn}$ are injected into the last $L_2$ (usually $L_2=4$) layers of a ViT encoder and processed jointly with patch tokens, yielding updated queries $\mathbf{Q}^\mathcal{S}$ which can be used to produce segmentation predictions $\{(\mathbf{c}_i, \mathbf{m}_i)\}_{i=1}^K$. Despite its simplicity, EoMT performs competitively with complex frameworks while greatly improving efficiency.

Inspired by this result, we explore a similar simplification for video segmentation, where inference speed is even more critical. Our hypothesis is that a strong ViT can handle not only segmentation but also temporal association within a unified \emph{encoder-only} architecture, removing the need for explicit tracking modules.
To verify this, starting from the state-of-the-art CAVIS model, we first replace its heavy segmenter with EoMT, and then we progressively remove video-specific components to evaluate whether the encoder can also learn to conduct temporal association.

\PAR{Replacing the Segmenter.} 
In the current state-of-the-art video segmentation models, such as CAVIS~\cite{lee2025cavis}, the segmenter $\mathcal{S}$ is composed of an inefficient ViT-Adapter~\cite{chen2023vitadapter} and a complex and resource-intensive Mask2Former~\cite{cheng2022mask2former} pixel decoder and Transformer decoder. We replace the entire segmenter with EoMT~\cite{kerssies2025eomt}, which integrates query tokens directly into the ViT and predicts object representations without specialized components. This greatly simplifies the pipeline and it is expected to consistently improve inference speed, similar to the original findings for EoMT~\cite{kerssies2025eomt}.

\PAR{Removing Context‐Aware Features.}
The context-aware features in CAVIS~\cite{lee2025cavis} explicitly encode information from the spatial neighborhood of each instance to stabilize predictions under appearance changes or occlusion. 
Extracting these features requires convolutional filtering over high-resolution features, repeated for every query in all frames of a video, making it inefficient.
We hypothesize that the auxiliary context added by these features is not strictly necessary when leveraging a strong pre-trained ViT, as its features are already fine-grained enough to be easily fine-tuned to capture specific object identity and to maintain stability under appearance changes or occlusion. %

\PAR{Removing Re‐identification Layers.}  
While effective, re-identification layers add complexity at both inference and training time, where the associated contrastive losses are memory-intensive and slow to optimize. 
We hypothesize that with large-scale pre-training, the features of the ViT encoder already contain rich instance-level information. Since the segmentation queries explicitly cross-attend to these features, they effectively inherit this instance-discriminative knowledge and preserve it across frames. Therefore, we can eliminate these layers, to not only simplify the whole pipeline but also make training more affordable.  %

\subsection{VidEoMT}
\label{sec:videomt}
After the previously described simplifications, the model consists of EoMT~\cite{kerssies2025eomt} as the segmenter combined with a simplified tracker $\mathcal{T}$. 
The tracker ensures that a given object is represented by the same query index across frames, preserving temporal consistency. However, this comes at the cost of considerable architectural complexity and significant computational overhead.

 We hypothesize that strong pre-training, \eg, with DINOv2~\cite{oquab2023dinov2}, already equips the ViT encoder with representations strong enough to enable temporal association within the encoder itself with only minimal changes, without the need for specialized tracking components. Hence, we move away from the conventional decoupling of segmenter and tracker and adopt a unified encoder-only design.

Enabling temporal modeling within an encoder-only framework presents two key challenges: (i) effectively integrating information from the previous frame to maintain temporal continuity, and (ii) preserving the model’s ability to detect and recognize newly appearing objects. To address the first challenge, we introduce a \textit{query propagation} mechanism that carries object-level information across frames, enabling temporal continuity within the encoder. To handle the second challenge, we propose a lightweight \textit{query fusion} strategy that combines propagated queries with learnable ones, allowing the model to better detect newly appearing objects. The resulting model, which we name \textit{Video Encoder-only Mask Transformer} (VidEoMT), performs temporal association without specialized tracking components, as visualized in \cref{fig:arch}. 

\PAR{Query Propagation.}
When the tracker is entirely removed, the model reduces to a purely image-level EoMT architecture that processes each frame independently. 
In this case, queries $\mathbf{Q}^\mathcal{S}$ are the model's output, and there are no queries $\mathbf{Q}^\mathcal{T}$, as there is no longer a tracker.

We reintroduce temporal modeling through query propagation. At timestep $t=0$, we follow the standard EoMT setup and feed learnable queries $\mathbf{Q}^\textrm{lrn}$ into the last $L_2$ layers of the ViT to produce object query embeddings $\mathbf{Q}^\mathcal{S}_0$ and the corresponding segmentation predictions. At subsequent timesteps $t>0$, instead of reusing the learnable queries, we use the track queries, \ie, the propagated queries from the previous frame $\mathbf{Q}^\mathcal{S}_{t-1}$, as input to the last $L_2$ layers of the ViT. During these timesteps, the segmentation procedure remains identical to that of EoMT, the only difference is that the propagated queries replace the learnable ones.

This strategy enables information to flow across time without additional computational cost per frame, allowing for temporal consistency across frames. However, since the ViT only receives information from the previous frame, the influence of the learnable queries $\mathbf{Q}^{\textrm{lrn}}$ gradually diminishes, causing the model to lose the ability to recognize newly appearing objects in the video.

\PAR{Query Fusion.}  
To address this limitation, we introduce \textit{query fusion}, illustrated in Figure~\ref{fig:arch}. In this design, queries from the previous frame $\mathbf{Q}^\mathcal{S}_{t-1}$ are first transformed by a lightweight linear layer and then combined with the original learned queries $\mathbf{Q}^\textrm{lrn}$ through element-wise addition:
\begin{equation}
    \mathbf{Q}_{t}^\mathcal{F} = \texttt{Linear}\!\left(\mathbf{Q}^\mathcal{S}_{t-1}\right) + \mathbf{Q}^\textrm{lrn}.
\end{equation}  
The element-wise addition is possible because the supervision strategy guarantees that the query order remains the same across frames.
This fusion ensures that the model has access to temporal context from the past through $\mathbf{Q}^\mathcal{S}_{t-1}$, as well as learnable queries $\mathbf{Q}^\textrm{lrn}$ to enable adaptability to new objects. By balancing temporal context propagation and adaptability to new objects, query fusion allows our encoder-only framework to conduct accurate object tracking with negligible additional architectural complexity.

\PAR{Training.} VidEoMT is trained using the same objective function as Mask2Former~\cite{cheng2022mask2former}. We use the cross-entropy loss for classification and the binary cross-entropy and Dice losses for segmentation predictions. To ensure temporally consistent supervision, we follow the ground-truth matching strategy of DVIS++~\cite{zhang2025dvis++}. Here, a ground-truth object is only matched to a query in the frame where the object first appears. In the remaining frames, the ground-truth object stays matched to this query, ensuring temporal consistency.

\section{Experiments}
\label{sec:exp}

\PARbegin{Datasets and Evaluation Metrics.} We evaluate VidEoMT on six major benchmarks for video segmentation: OVIS~\cite{qi2022occluded} and YouTube-VIS 2019, 2021, and 2022~\cite{yang2019video} for Video Instance Segmentation (VIS), VIPSeg~\cite{miao2022large} for Video Panoptic Segmentation (VPS), and VSPW~\cite{miao2021vspw} for Video Semantic Segmentation (VSS). We adopt Average Precision (AP) and Average Recall (AR) metrics~\cite{yang2019video} for VIS, Video Panoptic Quality (VPQ)~\cite{kim2020video} and Segmentation and Tracking Quality (STQ)~\cite{weber2021step} for VPS, and mean IoU (mIoU) and Video Consistency (mVC)~\cite{miao2021vspw} for VSS. 

\PAR{Implementation Details.} Similar to the state-of-the-art models CAVIS~\cite{lee2025cavis} and DVIS-DAQ~\cite{zhou2024dvisdaq}, we use a DINOv2-pretrained ViT~\cite{oquab2023dinov2} as the default backbone of VidEoMT. We adopt a batch size of 8 with 5 frames as a temporal window, using mixed precision and the AdamW optimizer~\cite{loshchilov2019adamw} with a learning rate of $10^{-4}$. Following EoMT~\cite{kerssies2025eomt}, we apply layer-wise learning rate decay (LLRD)~\cite{devlin2018bert} with a factor of 0.6 and a polynomial learning rate decay with a power of 0.9. The number of iterations and training video resolutions follow the settings of CAVIS~\cite{lee2025cavis} for fair comparison. We refer to the supplementary material for additional implementation details.

To assess the computational efficiency, we measure both FPS and FLOPs. FPS is reported as the average number of video frames processed per second on the validation set with a batch size of 1, evaluated on an NVIDIA H100 GPU with FlashAttention v2~\cite{dao2024flashattention2} and \texttt{torch.compile}~\cite{ansel2024pytorch2} (default settings) enabled. FLOPs are calculated using \textit{fvcore}~\cite{meta2023fvcore}, averaging over all images in the validation set. We report the results in GFLOPs, \ie, FLOPs $\times 10^9$.

\section{Results}

\subsection{Main Results}
\label{sec:results}

\PARbegin{From CAVIS to VidEoMT.}%
In~\cref{tab:steps_cavis}, we report a stepwise transformation from state-of-the-art video segmentation method CAVIS~\cite{lee2025cavis} to our proposed VidEoMT. We gradually remove specialized tracking components to obtain the lightweight EoMT baseline, and we then introduce modifications to EoMT to support tracking. For details of the architectures at different steps, see the supplementary.

In step \tablestep{1}, we find that replacing the segmenter with EoMT~\cite{kerssies2025eomt} improves FPS by almost 3$\times$, while AP drops by only 0.8. In steps \tablestep{2}--\tablestep{3}, we observe that removing \textit{context-aware features} and the \textit{re-identification layers} further increases speed by 1.8$\times$ to 74 FPS, with almost no impact on accuracy. These results demonstrate that the DINOv2 ViT encoder can take over the functionality of these components without degrading performance. In step \tablestep{4}, we note that the elimination of the tracker, which results in the naive, per-frame application of EoMT, yields a speedup of more than \textbf{10$\times$} to 162 FPS compared to CAVIS's 15 FPS, but also leads to a substantial 7.6 AP drop. Interestingly, though, even without any tracking modules and just relying on the queries, the model still retains reasonable accuracy. This shows that EoMT can learn to output objects in a somewhat consistent order across frames, despite processing them independently without temporal interaction.

\begin{table}[t]
    \centering
    \footnotesize
    \setlength{\tabcolsep}{1pt}
    \begin{tabularx}{\linewidth}{c l YYYY}
        \toprule
        Step & Method & AP & Params & GFLOPs & FPS \\
        
        \midrule

        \tablestep{0} & 
        CAVIS~\cite{lee2025cavis} & 
        68.9 & 
        358M & 
        838 &  
        15 \\
        
        \tablestep{1} & 
        \downrightarrow\; w/ EoMT as \textit{Segmenter} & 
        68.1 & 
        328M & 
        699 & 
        42 \\
        
        \tablestep{2} & 
        \downrightarrow\; w/o Context-aware Features & 
        68.4 & 
        327M & 
        581 & 
        72 \\
        
        \tablestep{3} & 
        \downrightarrow\; w/o Re-identification Layers & 
        68.0 & 
        326M & 
        580 & 
        74 \\
        
        \tablestep{4} & 
        \downrightarrow\; w/o Tracker = EoMT & 
        61.3 & 
        316M & 
        565 & 
        162 \\
        
        \midrule
        
        -- & 
        EoMT~\cite{kerssies2025eomt} & 
        61.3 & 
        316M & 
        565 & 
        162 \\
        
        \tablestep{5} & 
        \downrightarrow\; w/ Query Propagation & 
        63.9 & 
        316M & 
        565 & 
        162 \\
        
        \tablestep{6} & 
        \downrightarrow\; w/ Query Fusion = \textbf{VidEoMT} & 
        68.6 & 
        318M & 
        566 & 
        160 \\
        
        \bottomrule
    \end{tabularx}

    \caption{\textbf{From CAVIS to VidEoMT.} Stepwise removal of CAVIS modules toward EoMT, and modifications extending it to our VidEoMT. 
    Evaluated on YouTube-VIS 2019 \textit{val}~\cite{yang2019video}.}
    \label{tab:steps_cavis}
\end{table}

 \begin{table*}[t!]
     \centering
     \footnotesize
     \renewcommand{\tabcolsep}{3pt}
     \begin{tabularx}{\linewidth}
     {lll c YYYYY c YYYYY}
     \toprule
     \multirow{2}[2]{*}{Method} & \multirow{2}[2]{*}{Backbone} & \multirow{2}[2]{*}{Pre-training} && 
     \multicolumn{5}{c}{YouTube-VIS 2019 \textit{val}~\cite{yang2019video}} &&
     \multicolumn{5}{c}{YouTube-VIS 2021 \textit{val}~\cite{yang2019video}} \\
     \cmidrule{5-9} \cmidrule{11-15}
     &&&&
     AP & AP\textsubscript{75} & AR\textsubscript{10} & GFLOPs & FPS &&
     AP & AP\textsubscript{75} & AR\textsubscript{10} & GFLOPs & FPS \\
     \midrule
     MinVIS~\cite{huang2022minvis} & Swin-L & IN21K &&
     61.6 & 68.6 & 66.6 & 401 & 29 && %
     55.3 & 62.0 & 60.8 & 255 & 30 %
     \\
     DVIS~\cite{zhang2023dvis} & Swin-L & IN21K &&
     63.9 & 70.4 & 69.0 & 411 & 23 && %
     58.7 & 66.6 & 64.6 & 405 & 24 %
     \\
     DVIS-DAQ~\cite{zhou2024dvisdaq} & Swin-L & IN21K &&
     65.7 & 73.6 & 70.7 & 415 & 13 && %
     61.1 & 68.2 & 66.6 & 410 & 11  %
     \\
     DVIS++~\cite{zhang2025dvis++} & ViT-L & DINOv2 &&
     67.7 & 75.3 & 73.7 & 846 & 18  && %
     62.3 & 70.2 & 68.0 & 830 & 17 %
     \\
     DVIS-DAQ~\cite{zhou2024dvisdaq} & ViT-L & DINOv2 &&
     68.3 & 76.1 & 73.5 & 851 & 10 && %
     62.4 & 70.8 & 68.0 & 836 & 10 %
     \\
     CAVIS~\cite{lee2025cavis} & ViT-L & DINOv2 &&
     68.9 & 76.2 & 73.6 & 838 & 15 && %
     64.6 & 72.5 & 69.3 & 824 & 15 %
     \\
     \midrule
     \textbf{VidEoMT} & ViT-L & DINOv2 &&
     68.6 & 75.6 & 73.9 & 566 & 160 && %
     63.1 & 69.3 & 68.1 & 560 & 160 %
     \\
     \bottomrule
       
     \end{tabularx}
     \caption{\textbf{Online VIS on YouTube-VIS 2019 and 2021.}}
     \label{tab:sota_comparison_ytvis19_ytvis21}
 \end{table*}

\begin{table*}[t!]
    \centering
    \footnotesize
    \renewcommand{\tabcolsep}{3pt}
    \begin{tabularx}{\linewidth}
    {lll c YYYYY c YYYYY}
    \toprule
    \multirow{2}[2]{*}{Method} & \multirow{2}[2]{*}{Backbone} & \multirow{2}[2]{*}{Pre-training} && 
    \multicolumn{5}{c}{YouTube-VIS 2022 \textit{val}~\cite{yang2019video}} &&
    \multicolumn{5}{c}{OVIS \textit{val}~\cite{qi2022occluded}} \\
    \cmidrule{5-9} \cmidrule{11-15}
    &&&&
    AP$^{\text{L}}$ & AP$^{\text{L}}_{\text{75}}$ & AR$^{\text{L}}_{\text{10}}$ & GFLOPs & FPS &&
    AP & AP\textsubscript{75} & AR\textsubscript{10} & GFLOPs & FPS  \\
    \midrule
    MinVIS~\cite{huang2022minvis} & Swin-L & IN21K &&
    33.1 & 33.7 & 36.6 & 224 & 31 && %
    39.4 & 41.3 & 43.3 & 408 & 30  %
    \\

    DVIS~\cite{zhang2023dvis} & Swin-L & IN21K &&
    39.9 & 42.6 & 44.9 & 401 & 23 && %
    45.9 & 48.3 & 51.5 & 419 & 24 %
    \\
    DVIS-DAQ~\cite{zhou2024dvisdaq} & Swin-L & IN21K &&
    -- & -- & -- & -- & -- && %
    49.5 & 51.7 & 54.9 & 423 & 12 %
    \\
    DVIS++~\cite{zhang2025dvis++} & ViT-L & DINOv2 &&
    37.5 & 39.4 & 43.5 & 820 & 18 && %
    49.6 & 55.0 & 54.6 & 868 & 17  %
    \\
    CAVIS~\cite{lee2025cavis} & ViT-L & DINOv2 &&
    39.5 & 40.5 & 44.9 & 815 & 15 && %
    53.2 & 59.1 & 58.2 & 863 & 15 %
    \\
    DVIS-DAQ~\cite{zhou2024dvisdaq}$^\dagger$ & ViT-L & DINOv2 &&
    42.0 & 43.0 & 48.4 & 826 & 10 && %
    54.3 & 60.2 & 59.8 & 1173 & 8  %
    \\
    \midrule
    \textbf{VidEoMT}$^\dagger$ & ViT-L & DINOv2 &&
    42.6 & 46.1 & 48.1 & 557 & 161 &&
    52.5 & 57.2 & 57.5 & 934 & 115  %
    \\

    \bottomrule
      
    \end{tabularx}
    \caption{\textbf{Online VIS on YouTube-VIS 2022 and OVIS.} $^\dagger$Input resolution of 544 (shortest image side) for OVIS, default for others.}
    \label{tab:sota_comparison_ytvis22_ovis}
\end{table*}

Applying query propagation in step~\tablestep{5} is necessary to 
introduce temporal modeling in EoMT, improving the AP by +2.6 without increasing the computational cost. However, we find that the model struggles with identifying newly appearing objects over time.
In the final step~\tablestep{6}, we show that query fusion allows VidEoMT to recover nearly all of the original accuracy, while achieving a speedup of more than \textbf{10$\times$} compared to CAVIS.
Notably, the gain in inference speed is much larger than in FLOPs. This is the case because VidEoMT almost purely consists of a plain ViT. As such, it can better leverage dedicated hardware and software optimizations for the Transformer architecture without being bottlenecked by inefficient specialized modules~\cite{kerssies2025eomt}.

Overall, these results show that VidEoMT achieves an excellent balance between accuracy and efficiency, as heavy modules in CAVIS can be safely removed, while our encoder-only framework effectively restores performance with negligible computational cost.
Moreover, these results confirm our hypothesis that a VFM-pretrained ViT can be trained to conduct both segmentation and tracking within the same encoder, without complex tracking modules.

\subsection{Comparison with State-of-the-Art Models}
\label{sec:comparison_sota}

\PARbegin{Video Instance Segmentation (VIS).}
We first compare VidEoMT with state-of-the-art VIS models across four datasets. The results, reported in \cref{tab:sota_comparison_ytvis19_ytvis21,tab:sota_comparison_ytvis22_ovis}, demonstrate that VidEoMT consistently outperforms DVIS~\cite{zhang2023dvis} and DVIS++~\cite{zhang2025dvis++}, while being 5$\times$--8$\times$ faster. Compared to DVIS-DAQ~\cite{zhou2024dvisdaq}, VidEoMT is over 14$\times$ faster while achieving higher accuracy on all benchmarks except OVIS, where the gap is within 2 AP points. Similarly, VidEoMT surpasses CAVIS on YouTube-VIS 2022, and achieves comparable accuracy on YouTube-VIS 2019 and OVIS, and remains within 2 AP on YouTube-VIS 2021, while being 7$\times$ to more than 10$\times$ faster. Finally, we note that VidEoMT is also both faster and more accurate than MinVIS~\cite{huang2022minvis}, which was specifically designed for efficiency and simplicity.  
Overall, VidEoMT demonstrates a significantly superior trade-off between accuracy and efficiency compared to existing approaches.

\begin{table}[t]
    \centering
    \footnotesize
    \renewcommand{\tabcolsep}{2pt}
    \begin{tabularx}{\linewidth}
    {lll c YYcY}
    \toprule
    \multirow{2}[2]{*}{Method} & \multirow{2}[2]{*}{Backbone} & \multirow{2}[2]{*}{Pre-training} && 
    \multicolumn{4}{c}{VIPSeg \textit{val}~\cite{miao2022large}} \\
    \cmidrule{5-8} 
    &&&&
    VPQ & STQ & GFLOPs & FPS \\
    \midrule
    DVIS~\cite{zhang2023dvis} & Swin-L & IN21K && 54.7 & 47.7 & 879 & 20  %
    \\
    DVIS++~\cite{zhang2025dvis++} & ViT-L & DINOv2 && 56.0 & 49.8 & 2290 & 13  %
    \\
    CAVIS~\cite{lee2025cavis} & ViT-L & DINOv2 && 56.9 & 51.0 & 2612 & 10  %
    \\
    DVIS-DAQ~\cite{zhou2024dvisdaq} & ViT-L & DINOv2 && 57.4 & 52.0 & 2315 & 4  %
    \\
    \midrule
    \textbf{VidEoMT} & ViT-L & DINOv2 && 55.2 & 48.9 & 1897 & 75 %
    \\
    \bottomrule
     
    \end{tabularx}
    \caption{\textbf{Online VPS on VIPSeg.}}
    \label{tab:sota_comparison_vipseg}
\end{table}

\begin{table}[t!]
    \centering
    \footnotesize
    \renewcommand{\tabcolsep}{2pt}
    \begin{tabularx}{\linewidth}
    {lll c cYcY}
    \toprule
    \multirow{2}[2]{*}{Method} & \multirow{2}[2]{*}{Backbone} & \multirow{2}[2]{*}{Pre-training} && 
    \multicolumn{4}{c}{VSPW \textit{val}~\cite{miao2021vspw}} \\
    \cmidrule{5-8} 
    &&&&
    mVC\textsubscript{16} & mIoU & GFLOPs & FPS \\
    \midrule
    DVIS~\cite{zhang2023dvis} & Swin-L & IN21K && 94.3 & 61.3 & 879 & 22%
    \\
    DVIS++~\cite{zhang2025dvis++} & ViT-L & DINOv2 && 94.2 & 62.8 & 2290 & 13  %
    \\
    \midrule
    \textbf{VidEoMT} & ViT-L & DINOv2 && 95.0 & 64.9 & 1909 & 73  %
    \\
  
    \bottomrule
      
    \end{tabularx}
    \caption{\textbf{Online VSS on VSPW.}}
    \label{tab:sota_comparison_vspw}
\end{table}

\PAR{Video Panoptic Segmentation (VPS).} \cref{tab:sota_comparison_vipseg} compares VidEoMT with state-of-the-art methods for the VPS task on the VIPSeg benchmark. %
VidEoMT incurs only a minor VPQ drop compared to DVIS++ and CAVIS, while running 5$\times$--7$\times$ faster. 
Compared to DVIS-DAQ, which obtains the highest VPQ of 57.4 but runs at the lowest FPS of 4, VidEoMT sacrifices just 2.2 VPQ while delivering nearly 19$\times$ higher speed.
These results confirm that VidEoMT also provides a significantly better accuracy and efficiency balance for video panoptic segmentation.

\PAR{Video Semantic Segmentation (VSS).} \cref{tab:sota_comparison_vspw} compares VidEoMT with state-of-the-art VSS methods on the VSPW benchmark.
VidEoMT outperforms existing methods, improving the mIoU by +2.1 compared to DVIS++ while also achieving a higher temporal consistency with +0.8 mVC\textsubscript{16} and being more than 5$\times$ faster.
These results confirm the general applicability and strength of VidEoMT on yet another video segmentation task.

\subsection{Further Analyses}
\label{ablation}

\PARbegin{EoMT as a Segmenter.}
 In this work, we propose a unified architecture that performs video segmentation in an encoder-only fashion. Alternatively, one could consider augmenting EoMT with recent trackers.
In \cref{tab:alter_app_eomt_as_segmenter}, we compare VidEoMT to alternative approaches where EoMT is used as a segmenter and existing trackers are applied on top.
Compared to the best alternative approach, EoMT + CAVIS, VidEoMT achieves slightly better AP while being nearly $4 \times$ faster.
These results show that our VidEoMT is not only more streamlined but also considerably faster and even more accurate than alternative strategies.

\PAR{Query Propagation.}
In VidEoMT, we directly propagate object queries into the ViT encoder.
To verify that this is just as effective as propagating them into a separate decoder, we take a DINOv2 + ViT-Adapter encoder and Mask2Former decoder~\cite{cheng2022mask2former}, and apply query propagation into the decoder.
We evaluate two propagation variants: TrackFormer~\cite{meinhardt2022trackformer}, which was originally introduced for object tracking rather than video segmentation, and our query fusion approach that combines propagated queries with learned queries.
The results in \cref{tab:alter_app_query_propagation} show that our encoder-only approach achieves an AP score comparable to the encoder--decoder design, validating the effectiveness of the proposed encoder-only architecture.
Furthermore, our query fusion strategy slightly improves AP over TrackFormer~\cite{meinhardt2022trackformer} while also being significantly faster.
Overall, VidEoMT is just as effective but considerably faster than both alternative approaches. See the supplementary material for additional experiments on temporal propagation and more details on these alternative methods.

\PAR{Impact of Pre-training.}
In this work, we hypothesize that large-scale pre-training with VFMs enables the ViT encoder in VidEoMT to take over the functionalities of specialized components.
To verify this hypothesis, in \cref{tab:pretraining}, we compare the performance of VidEoMT and CAVIS using large-scale pre-training with DINOv3~\cite{simeoni2025dinov3}, DINOv2~\cite{oquab2023dinov2}, and EVA-02~\cite{fang2024eva02}, as well as medium-scale ImageNet-21K and small-scale ImageNet-1K pre-training~\cite{touvron2022deit, deng2009imagenet}. We observe that, with strong pre-training (DINOv2, DINOv3, EVA-02), VidEoMT attains performance comparable to CAVIS, while the performance gap increases in favor of CAVIS as the pre-training scale decreases. 
Notably, DINOv3 offers only marginal improvements over DINOv2, which we attribute to DINOv3 being designed to be kept frozen rather than fine-tuned. Nevertheless, VidEoMT remains highly effective under DINOv3 pre-training. Overall, these results support our hypothesis that large-scale pre-training is necessary to unleash the potential of VidEoMT.
While EoMT~\cite{kerssies2025eomt} showed this effect for image segmentation, our results demonstrate that large-scale pre-training 
also enables the ViT encoder to take over the functionalities of specialized video segmentation components.

\begin{table}[t]
    \centering
    \footnotesize
    \renewcommand{\tabcolsep}{2.5pt}
    \begin{tabularx}{\linewidth}
    {l c l c YYYcc}

    \toprule

    Segmenter && Tracker && AP & AP\textsubscript{75} & AR\textsubscript{10} & GFLOPs & FPS  \\

    \midrule

    EoMT && CAVIS~\cite{lee2025cavis} 
    && 
    68.1 & 76.3 & 73.6 & 699 & 42 \\

    EoMT && DVIS++~\cite{zhang2025dvis++} 
    && 
    67.0 & 73.8 & 73.2 & 683 & 69  \\

    EoMT && DVIS-DAQ~\cite{zhou2024dvisdaq} 
    && 
    67.3 & 74.8 & 73.4 & 703 & 28 \\

    \midrule

    \textbf{VidEoMT} && -- && 
    68.6 & 75.6 & 73.9 & 566 & 160  \\

    \bottomrule
    \end{tabularx}
    \caption{\textbf{Alternative approaches: EoMT as a segmenter.} Comparison of EoMT equipped with state-of-the-art trackers and our proposed VidEoMT. Evaluated on YouTube-VIS 2019 \textit{val}.}
    \label{tab:alter_app_eomt_as_segmenter}
\end{table}

\begin{table}[t]
    \centering
    \footnotesize
    \renewcommand{\tabcolsep}{3pt}
    \begin{tabularx}{\linewidth}
    {l c l c l c ccc}

    \toprule

    Encoder && Decoder && Temporal Modeling &&
    AP & GFLOPs & FPS \\

    \midrule
    
    ViT-Adapter && M2F~\cite{cheng2022mask2former} 
    && TrackFormer~\cite{meinhardt2022trackformer} 
    &&
    67.8 & 739 & 22 \\

    ViT-Adapter && M2F~\cite{cheng2022mask2former} 
    && Query Fusion &&
    68.0 & 718 & 32  \\

    \midrule

    \textbf{VidEoMT} && -- && Query Fusion && 
    68.6 & 566 & 160 \\

    \bottomrule
    \end{tabularx}
    \caption{\textbf{Alternative approaches: Query propagation in the decoder.} Comparison of ViT-Adapter + Mask2Former (M2F) equipped with TrackFormer or our query fusion strategy and the proposed VidEoMT. All methods use a ViT-L backbone with DINOv2 pre-training. Evaluated on YouTube-VIS 2019 \textit{val}.}
    \label{tab:alter_app_query_propagation}
\end{table}

\begin{table}[th]
    \centering
    \footnotesize
    \renewcommand{\tabcolsep}{2pt}
    \begin{tabularx}{\linewidth}{l Y YYY}

    \toprule

    Model & 
    Pre-training &
    AP & 
    GFLOPs & 
    FPS \\

    \midrule

    CAVIS & \multirow{2}{*}{DINOv3} &
    \phantom{\qdecnew{0.0}}68.8\phantom{\qdecnew{0.0}} & 
    838 & 
    13 
    \\
    \textbf{VidEoMT} & &
    \phantom{\qdecnew{0.0}}68.9\qincnew{0.1} & 
    566 & 
    133 
    \\
    
    \midrule

    CAVIS & \multirow{2}{*}{DINOv2} &
    \phantom{\qdecnew{0.0}}68.9\phantom{\qdecnew{0.0}} & 
    838 & 
    15 
    \\
    \textbf{VidEoMT} & &
    \phantom{\qdecnew{0.0}}68.6\qdecnew{0.3} & 
    566 & 
    160 
    \\
    
    \midrule

    CAVIS & \multirow{2}{*}{EVA-02} &
    \phantom{\qdecnew{0.0}}68.0\phantom{\qdecnew{0.0}} & 
    835 & 
    12 
    \\
    \textbf{VidEoMT} & &
    \phantom{\qdecnew{0.0}}67.8\qdecnew{0.2} & 
    564 & 
    119 
    \\
    
    \midrule

    CAVIS & \multirow{2}{*}{IN21K} &
    \phantom{\qdecnew{0.0}}62.2\phantom{\qdecnew{0.0}} & 
    838 & 
    15 
    \\
    \textbf{VidEoMT} & &
    \phantom{\qdecnew{0.0}}60.8\qdecnew{1.4} & 
    566 & 
    160 
    \\

    \midrule

    CAVIS & \multirow{2}{*}{IN1K} &
    \phantom{\qdecnew{0.0}}59.4\phantom{\qdecnew{0.0}} & 
    838 & 
    15 
    \\
    \textbf{VidEoMT} & &
    \phantom{\qdecnew{0.0}}56.7\qdecnew{2.7} & 
    566 & 
    160 
    \\

    \bottomrule
    \end{tabularx}
    \caption{\textbf{Impact of pre-training.} VidEoMT performs better with larger-scale pre-training. Evaluated on YouTube-VIS 2019 \textit{val}.}
    \label{tab:pretraining}
\end{table}

\begin{table}[th]
    \centering
    \footnotesize
    \renewcommand{\tabcolsep}{2.2pt}
    \begin{tabularx}{\linewidth}{l Y YYYY}

    \toprule

    Method & Size & AP & Params & GFLOPs & FPS \\

    \midrule
    
    CAVIS  & \multirow{2}{*}{L} &
    \phantom{\qdecnew{0.0}}68.9\phantom{\qdecnew{0.0}} & 358M & 838 & 15
    \\

    \textbf{VidEoMT} & &
    \phantom{\qdecnew{0.0}}68.6\qdecnew{0.3} & 318M & 566 & 160
    \\
    
    \midrule

    CAVIS & \multirow{2}{*}{B} &
    \phantom{\qdecnew{0.0}}59.5\phantom{\qdecnew{0.0}} & 131M & 390 & 18
    \\

    \textbf{VidEoMT} & &
    \phantom{\qdecnew{0.0}}58.2{\qdecnew{1.3}} & 95M & 182 & 251
    \\

    \midrule

    CAVIS  & \multirow{2}{*}{S} &
    \phantom{\qdecnew{0.0}}55.5\phantom{\qdecnew{0.0}} & 57M & 251 & 19
    \\
    
    \textbf{VidEoMT} & &
    \phantom{\qdecnew{0.0}}52.8{\qdecnew{2.7}} & 25M & 56 & 294
    \\

    \bottomrule
        
    \end{tabularx}
    \caption{\textbf{Impact of model size.} VidEoMT performs better as ViT~\cite{dosovitskiy2021vit} size increases. Evaluated on YouTube-VIS 2019 \textit{val}.} 
    \label{tab:model_size}
\end{table}

\PAR{Impact of Model Size.}
Similarly, we hypothesize that increased model size positively impacts the ViT's ability to conduct segmentation and tracking.
In \cref{tab:model_size}, we assess this by evaluating CAVIS and VidEoMT for ViT model sizes L, B, and S. 
The results show that the gap between the CAVIS baseline and VidEoMT decreases as model size increases, confirming our hypothesis.
Additionally, while there is a moderate performance gap between CAVIS and VidEoMT for smaller model sizes, VidEoMT with a large ViT-L backbone is still an order of magnitude faster than CAVIS with a small ViT-S backbone, while being much more accurate.
This further highlights the effectiveness of VidEoMT.

\section{Conclusion}
\label{sec:conclusion}

We have introduced VidEoMT, an \textit{encoder-only} video segmentation architecture that unifies segmentation and temporal association within a single ViT encoder. Through a step-by-step reduction of prior models, we showed that heavy specialized modules are no longer required. We replace them with a lightweight query propagation method, enhanced with an efficient query fusion mechanism. This design achieves an order-of-magnitude speedup while preserving or improving accuracy across multiple video segmentation benchmarks. Overall, our findings suggest that a sufficiently large and well-pretrained ViT can take over much of the functionality that was previously handled by complex downstream components in video segmentation. We hope that this work can serve as a foundation for applications with strict efficiency requirements.

\subsubsection*{Acknowledgments}
 This work was partly funded by the EU project MODI (grant no. 101076810), the KDT JU project EdgeAI (grant no. 101097300), and the BMFTR project WestAI (grant nos. 01IS22094D and 16IS22094D). The experiments utilized both the Dutch national infrastructure, supported by the SURF Cooperative under grant nos. EINF-14337, EINF-11307, and EINF-15136, and funded by the Dutch Research Council (NWO), and the computing resources provided by the Gauss Centre for Supercomputing e.V. through the John von Neumann Institute for Computing (NIC) on the GCS supercomputer JUWELS at Jülich Supercomputing Centre.

{
    \small
    \bibliographystyle{ieeenat_fullname}
    \bibliography{main}
}

\end{document}



\maketitlesupplementary

\normalsize

\renewcommand{\thetable}{\Alph{table}}
\setcounter{table}{0}
\renewcommand{\thefigure}{\Alph{figure}}
\setcounter{figure}{0}

\appendix
\section*{Appendix}
\label{appendix}
\paragraph{Table of contents}
\begin{itemize}
\item \S\ref{sec:supp:impl_details}: Implementation Details
\item \S\ref{sec:supp:add_ablations}: Additional Experiments
\item \S\ref{sec:supp:qual_results}: Qualitative Results
\item \S\ref{sec:supp:failure_analysis}: Failure Cases
\end{itemize}

\section{Implementation Details}
\label{sec:supp:impl_details}

\subsection{Training}
Following state-of-the-art models CAVIS~\citep{lee2025cavis}, DVIS-DAQ~\citep{zhou2024dvisdaq} and DVIS++~\citep{zhang2025dvis++}, we adopt a DINOv2-pretrained ViT~\citep{oquab2023dinov2,dosovitskiy2021vit} as the backbone of VidEoMT, and we train our model in two stages. In stage one, we train the model for image segmentation only. Concretely, we train on COCO~\cite{lin2014coco} instance segmentation and the target video segmentation dataset without applying any temporal supervision.

In the second stage, we introduce temporal modeling and fine-tune the model from stage one for video segmentation. Unlike CAVIS, DVIS-DAQ, and DVIS++, which freeze the DINOv2-initialized ViT encoder after stage one, we keep fine-tuning the ViT encoder for VidEoMT. We explore fine-tuning the ViT encoder for the CAVIS and DVIS++ baselines in Tabs.~1 and~2 as well, but find that the loss diverges or the memory increases beyond the GPUs' limits. For our VidEoMT, note that fine-tuning the encoder is necessary because our model is encoder-only, meaning that the encoder weights need to be optimized to allow the model to be trained for video segmentation.

\subsection{Evaluation} 

During evaluation, we process videos in a frame-by-frame fashion, as is required for online video segmentation. We evaluate efficiency in terms of FPS and GFLOPs. All metrics are measured on a single NVIDIA H100 GPU using PyTorch 2.7 and CUDA 12.6. We use a batch size of 1 frame to report mean values computed across all frames in the entire validation set.
\textbf{FPS} is measured using FlashAttention v2~\citep{dao2024flashattention2} and \texttt{torch.compile}~\citep{ansel2024pytorch2} with default settings and automatic mixed precision, after 100 warm-up iterations. 
\textbf{FLOPs} are measured with fvcore~\citep{meta2023fvcore}, and reported in GFLOPs (FLOPs $\times 10^{-9}$).

\captionsetup[subfigure]{labelformat=empty}
\begin{figure}[!htbp]
\centering
    \begin{subfigure}[b]{0.45\textwidth}
        \centering
        \includegraphics[width=\linewidth]{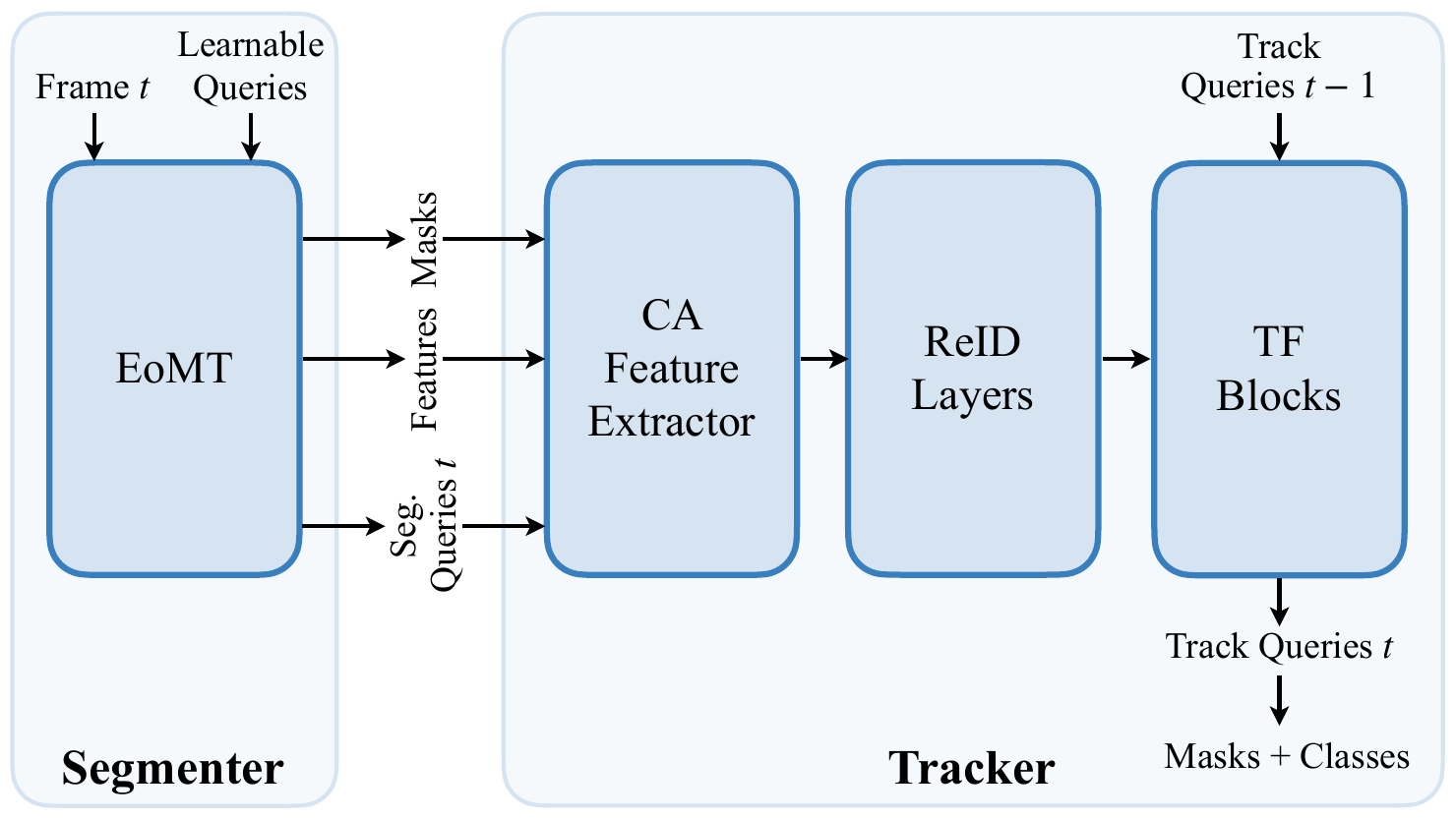}
        \caption{\textbf{Step (1): w/ EoMT as the Segmenter}}
        \label{fig:with_eomt_as_segmenter}
    \end{subfigure}
    \hfill
    \vspace{15pt}
        \begin{subfigure}[b]{0.45\textwidth}
        \centering
        \includegraphics[width=\linewidth]{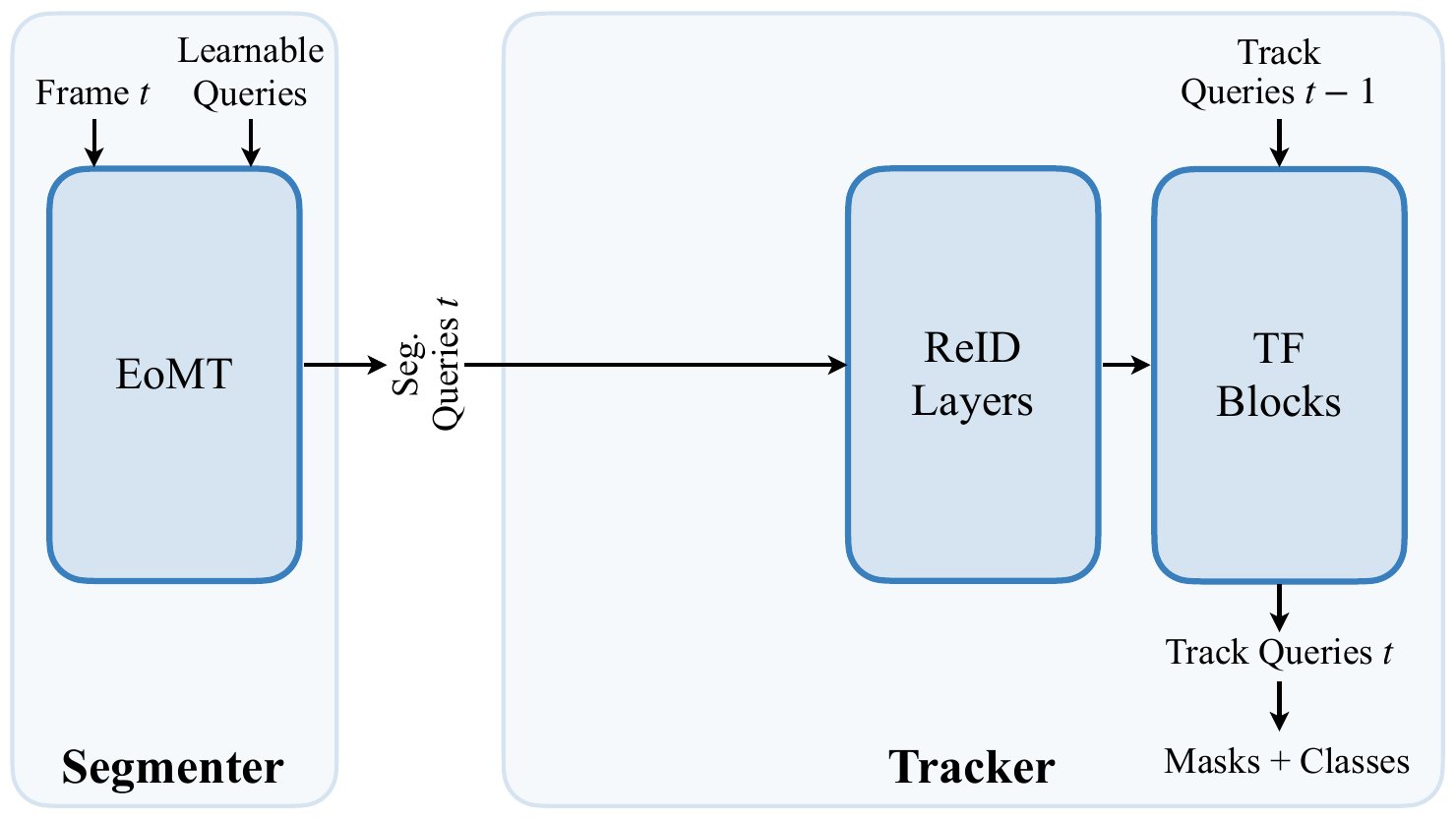}
        \caption{\textbf{Step (2): w/o Context-aware Features}}
        \label{fig:without_context_aware_features}
    \end{subfigure}
    \hfill
    \vspace{15pt}
    \begin{subfigure}[b]{0.45\textwidth}
        \centering
        \includegraphics[width=\linewidth]{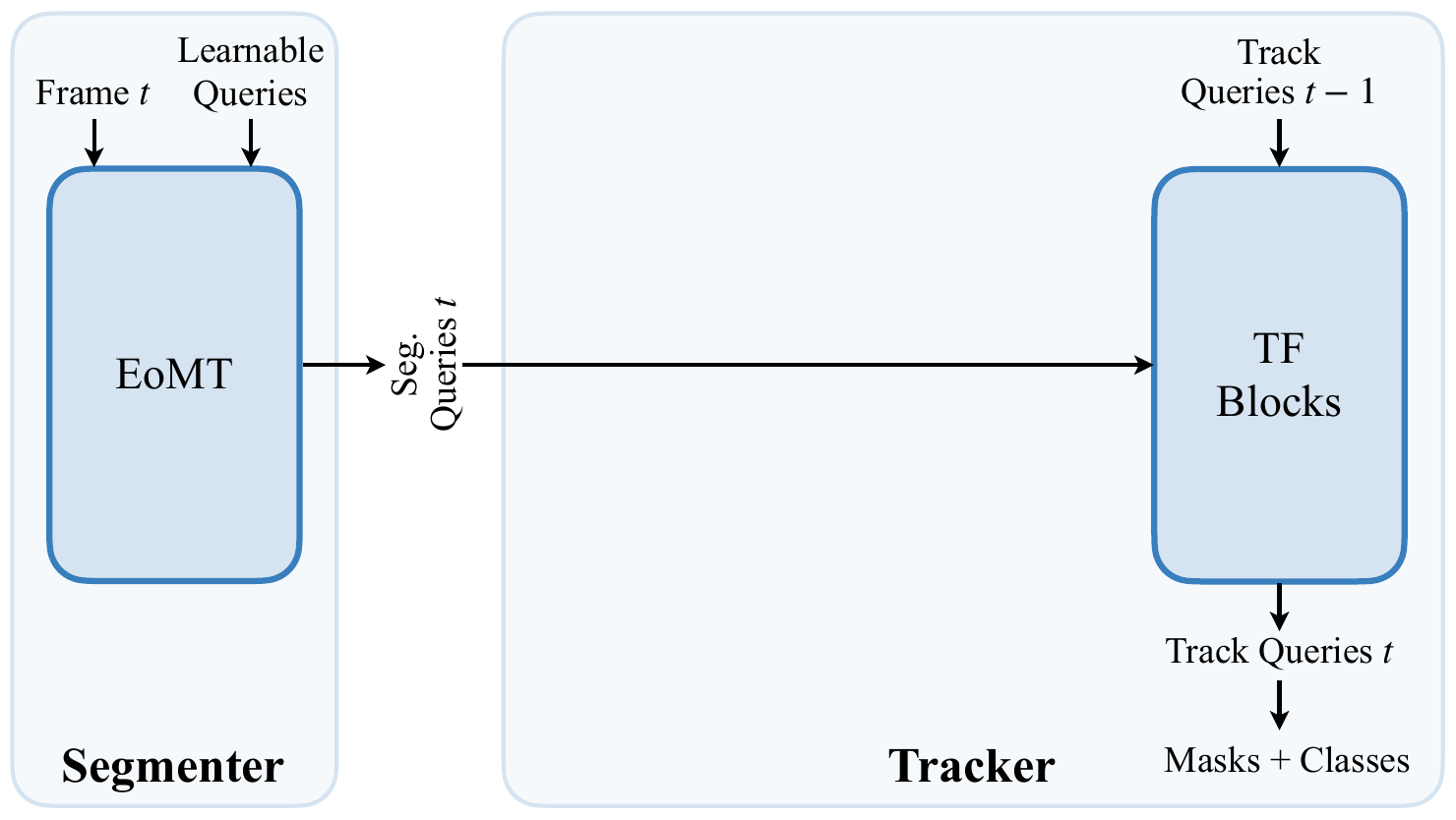}
        \caption{\textbf{Step (3): w/o Re-identification Layers}}
        \label{fig:without_reidentification_layers}
    \end{subfigure}
    \hfill
    \vspace{15pt}
    \begin{subfigure}[b]{0.45\textwidth}
        \centering
        \includegraphics[width=\linewidth]{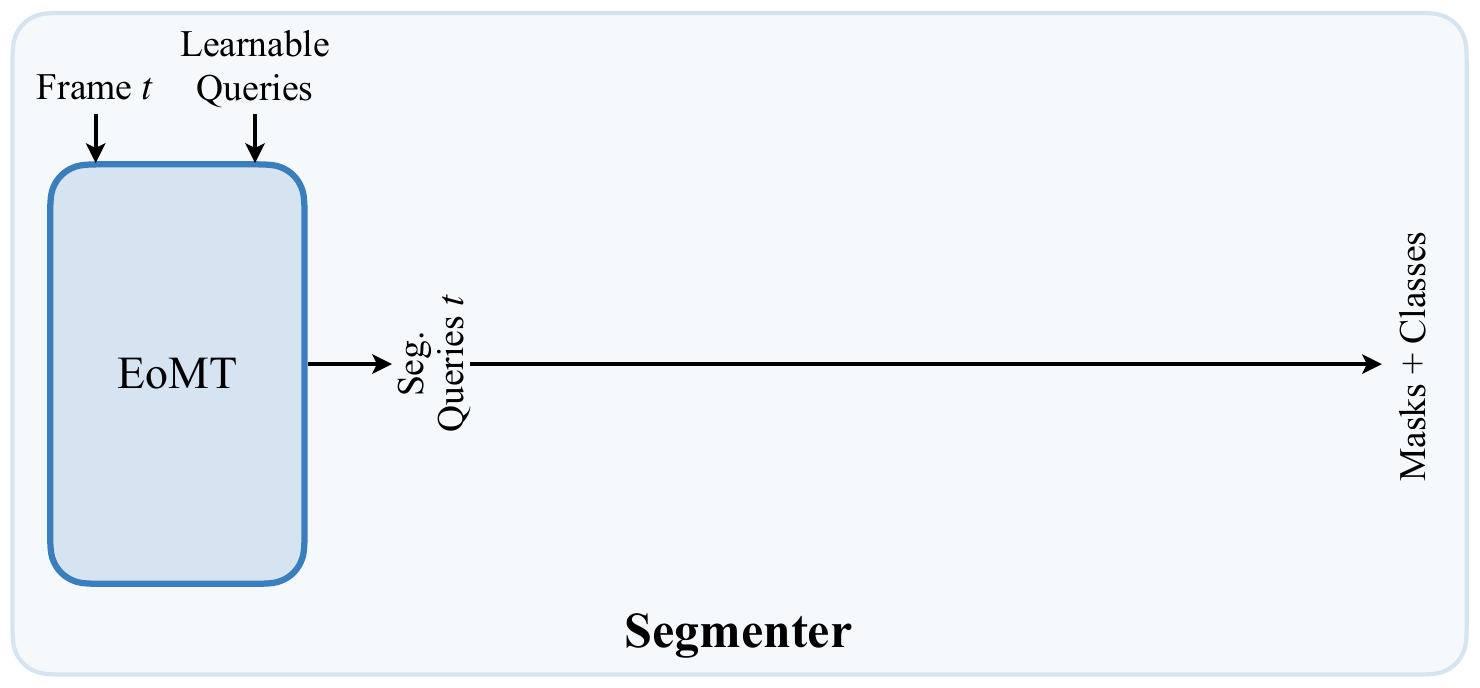}
        \caption{\textbf{Step (4): w/o Tracker Blocks}}
        \label{fig:without_tracker_blocks}
    \end{subfigure}
    \hfill
    \vspace{15pt}
\caption{\textbf{Removing specialized components.} This figure visualizes the step-by-step removal of complex,  specialized components from the CAVIS~\cite{lee2025cavis} model, as reported in the results in~Tab.~6 of the main manuscript.}
\label{app:fig:removing_steps}
\end{figure}

\subsection{Visualizations of Model Configurations}
\label{app:sub:Vis_model_confs}
In Sec.~3.3 and Tab.~1, we gradually remove specialized components from the state-of-the-art video segmentation model CAVIS~\cite{lee2025cavis}, which is visualized in Fig.~2 (left). 
To provide more details, we additionally illustrate the architectures at intermediate steps \tablestep{1} to \tablestep{4} in~\cref{app:fig:removing_steps}. 
In the first step, we replace CAVIS's original segmenter -- consisting of DINOv2~\cite{oquab2023dinov2}, ViT-Adapter~\cite{chen2023vitadapter}, and Mask2Former's pixel decoder and Transformer decoder~\cite{cheng2022mask2former} -- with EoMT~\cite{kerssies2025eomt}. 
In the second step, we remove the context-aware features module and directly forward the segmenter's output queries to the re-identification layers. 
In the third step, we also remove the re-identification layers, sending the segmenter's output queries directly to the tracker's Transformer blocks. 
Subsequently, in the fourth step, we discard the tracker altogether, and naively apply EoMT only on a per-frame basis. In this step, temporal association is then obtained in the simplest possible way: we assign all objects predicted from the same query across frames to the same track, without any additional post-hoc temporal matching. 

In step~\tablestep{5}, which we do not visualize here, we propagate queries by directly feeding the output segmentation queries from frame~$t-1$ into the encoder for frame~$t$. Finally, in step~\tablestep{6}, we introduce our query fusion design, where propagated queries are fused with learnable queries. The resulting architecture is visualized in Fig.~3 (right).

\subsection{Hyperparameters}
For step~\tablestep{0} in~Tab.~1, we report results using the official CAVIS~\cite{lee2025cavis} model weights, which we were able to reproduce. For all subsequent steps, we train the models using the same settings as CAVIS with respect to input size, number of iterations, batch size, and number of sampled frames. Specifically, we use a batch size of 8, train on 8 NVIDIA H100 GPUs, and sample 5 frames from a video clip. 
We train for 160k iterations on YouTube-VIS~\citep{yang2019video} (all versions) and OVIS~\citep{qi2022occluded}, for 40k iterations on VIPSeg~\citep{miao2022large}, and for 20k iterations on VSPW~\citep{miao2021vspw}. Additionally, all VidEoMT models use $N=200$ learnable queries with a feature dimension of $D=1024$.

For all experiments that use EoMT as the segmenter, as well as for all experiments with VidEoMT, we keep the optimization strategy identical to that of EoMT. Concretely, we use automatic mixed precision and the AdamW optimizer~\citep{loshchilov2019adamw} with a learning rate of $10^{-4}$.  We apply layer-wise learning rate decay (LLRD)~\citep{devlin2018bert} with a factor of 0.6 and polynomial learning rate decay with a power of 0.9. A two-stage linear warm-up strategy is used for all models, including the baselines. Specifically, we first warm up the randomly initialized parameters for 500 iterations while keeping the pre-trained parameters frozen. Then, after 500 iterations, we warm up the pre-trained parameters for 1000 iterations. In both stages, the initial learning rate is set to 0. 

To supervise our models, we adopt the same loss functions as Mask2Former~\citep{cheng2022mask2former}. Across all tasks and datasets, we use the cross-entropy (CE) loss for the classification predictions, and the binary cross-entropy (BCE) loss together with Dice loss for segmentation predictions. The total loss is a weighted sum of these components:
\begin{equation}
\mathcal{L}_{\textrm{tot}} = \lambda_{\textrm{bce}} \mathcal{L}_{\textrm{bce}} + \lambda_{\textrm{dice}} \mathcal{L}_{\textrm{dice}} + \lambda_{\textrm{ce}} \mathcal{L}_{\textrm{ce}}.
\end{equation}
where $\lambda_{\textrm{bce}}$, $\lambda_{\textrm{dice}}$, and $\lambda_{\textrm{ce}}$ are set to 5.0, 5.0, and 2.0, respectively, following Mask2Former~\citep{cheng2022mask2former}.

\subsection{Architectures of Alternative Approaches}

In~Tab.~7, we compare VidEoMT with an alternative encoder–decoder architecture that performs temporal modeling in the decoder, with two different temporal modeling approaches: our proposed query fusion and a TrackFormer-based design~\cite{meinhardt2022trackformer}. As the encoder, we use DINOv2 + ViT-Adapter~\citep{chen2023vitadapter,oquab2023dinov2}, and as the decoder we use a Transformer that follows the architecture of the Mask2Former Transformer decoder for segmentation.~\citep{cheng2022mask2former}. Concretely, we adopt the standard Mask2Former decoder with 9 layers, each composed of cross-attention, self-attention, and feed-forward blocks, operating with a hidden dimension of 256. To introduce temporal modeling, we feed the track queries and learnable queries into the decoder instead of the encoder's Transformer blocks, which we would do for VidEoMT. At the output of the decoder, the resulting queries are used to predict segmentation masks and classes in the same way as for VidEoMT. For query fusion, we adopt the same approach as described in Sec.~3.4.

The described encoder--decoder approach, which is much less efficient than the encoder-only VidEoMT method (see Tab.~7), somewhat resembles TrackFormer~\citep{meinhardt2022trackformer}, a method for bounding-box multi-object tracking (MOT). TrackFormer also applies temporal modeling by propagating queries into the decoder, but follows a more complex approach to do so. To assess the effectiveness of our query fusion approach compared to TrackFormer's approach, we therefore additionally implement TrackFormer's temporal modeling strategy in the encoder--decoder setting, while staying as close as possible to the original implementation.

Specifically, we make predictions for the first frame using a set of 400 learnable queries. Using these predictions, only the $N$ queries with a classification score $s > 0.8$ are kept and converted into \textit{track queries}. For the next frame, these track queries are concatenated with the 400 original learnable queries, which are then fed to the decoder for that frame. In subsequent frames, the decoder updates the propagated track queries such that they predict the masks for the same objects in the new frames. Again, newly detected queries with scores $s > 0.8$ are added as additional track queries, and non-maximum suppression (NMS) with an IoU threshold of $\sigma_{\text{NMS}} = 0.9$ is applied to remove near-duplicate predictions. Note that this NMS operation is the main reason for the TrackFormer approach's inefficiency compared to VidEoMT's query propagation mechanism. Finally, at each frame, track queries are removed if their score remains below $s < 0.8$ for five consecutive frames, indicating that the object they are tracking has disappeared from the scene.

\section{Additional Experiments}
\label{sec:supp:add_ablations}

\vspace{-10pt}
\phantom{x}

\begin{table}[t]
    \centering
    \footnotesize
    \renewcommand{\tabcolsep}{2.5pt}
    \begin{tabularx}{\linewidth}{l c YYY}

    \toprule

    \multirow{2}[2]{*}{Method} & & 
    \multicolumn{3}{c}{YouTube-VIS 2019 \textit{val}~\cite{yang2019video}} \\

    \cmidrule{3-5}

    & & AP & GFLOPs & FPS \\

    \midrule

    No propagation  & & 61.3 & 565 & 162 \\
    Propagation only  & & 63.9 & 565 & 162 \\
    Non-object reset  & & 67.8 & 565 & 157 \\
    TrackFormer & & 67.7 & 571 & 117 \\

    \midrule
    \textbf{Fusion}$\;\Rightarrow\;$ \textbf{VidEoMT} & & 68.6 & 566 & 160 \\

    \bottomrule
        
    \end{tabularx}
    \caption{\textbf{Query propagation methods.} Comparison of alternative strategies for temporal propagation.}
    \label{tab:query_update}
\end{table}

\subsection{Query Propagation Methods} 
VidEoMT propagates queries by fusing the learnable queries with the propagated track queries.
In \cref{tab:query_update}, we compare this approach with alternative methods to propagate queries.

We start with the \textit{no propagation} approach, the simplest variant, where the model receives only the learnable queries -- similar to EoMT -- but is fine-tuned for video segmentation. This setting performs the worst, as it lacks any form of explicit temporal modeling.

Next, in the \textit{propagation only} variant, we introduce temporal modeling by directly propagating the output queries from the previous frame into the current frame's encoder. This is step \tablestep{5} in Tab.~1. However, this approach struggles to detect new objects effectively, as the influence of the learnable queries diminishes over time.

\textit{Non-object reset} improves over this by replacing a propagated query with a learnable query if it did not predict an object in the previous frame, but this still underperforms the default fusion approach.

Finally, we evaluate the \textit{TrackFormer} approach~\citep{meinhardt2022trackformer} of only propagating queries for detected objects and introducing new learnable queries to detect new objects. 
This approach performs slightly worse than our \textit{fusion} approach, but most importantly it is considerably slower because it requires filtering out duplicate detections that should not be propagated.
Overall, these results demonstrate that our \textit{fusion} approach is the most accurate and efficient.

\subsection{Impact of Model Size}
In~Tab.~9 of the main manuscript, we report the impact of model size for both VidEoMT and CAVIS. In this section, in \cref{app:tab:model_size}, we additionally report the results of the EoMT + CAVIS combination, which we also visualize in~Fig.~1.
Compared to the alternative approach of extending EoMT with a CAVIS tracker, VidEoMT consistently performs better in terms of both efficiency and accuracy across all backbones. This highlights the effectiveness of VidEoMT over the more naive approach of extending EoMT with a state-of-the-art tracker.

\begin{table}[!t]
    \centering
    \footnotesize
    \renewcommand{\tabcolsep}{2.2pt}
    \begin{tabularx}{\linewidth}{l Y YYYY}

    \toprule

    Method & Size & AP & Params & GFLOPs & FPS \\

    \midrule
    
    CAVIS  && 68.9 & 358M & 838 & 15
    \\
      
    EoMT + CAVIS & L & 68.1 & 328M & 699 & 42
    \\

    \textbf{VidEoMT} & & 68.6 & 318M & 566 & 160
    \\
    
    \midrule

    CAVIS &  & 59.5 & 131M & 390 & 18
    \\ 
    EoMT + CAVIS & B & 57.4& 103M & 284 & 67
    \\

    \textbf{VidEoMT} & & 58.2 & 95M & 182 & 251
    \\

    \midrule

    CAVIS  & & 55.5 & 57M & 251 & 19
    \\
    EoMT + CAVIS & S & 50.3 & 34M & 150 & 93
    \\
    
    \textbf{VidEoMT} & & 52.8 & 25M & 56 & 294
    \\

    \bottomrule
        
    \end{tabularx}
    \caption{\textbf{Impact of model size.} VidEoMT performs better as ViT~\cite{dosovitskiy2021vit} size increases. Evaluated on YouTube-VIS 2019 \textit{val}.} 
    \label{app:tab:model_size}
\end{table}

\subsection{Impact of Pre-training}
In Tab.~8 of the main manuscript, we observe that DINOv3~\cite{simeoni2025dinov3} and EVA-02~\cite{fang2024eva02} are slower than DINOv2~\cite{oquab2023dinov2}, despite having a similar number of GFLOPs. Since both DINOv3 and EVA-02 use rotary positional embeddings (RoPE), we attribute a significant part of this slowdown to RoPE, as it introduces additional element-wise operations in the attention layers. When we disable RoPE in these models, we obtain faster models, confirming that RoPE is one of the main sources of the slowdown. Other implementation details may also play a secondary role.

\definecolor{caviscolor}{HTML}{397FBE}
\definecolor{videomtcolor}{HTML}{ff6d01}
\begin{figure}[t]
\centering
\begin{tikzpicture}
\begin{axis}[
    width=0.9\linewidth,
    height=6.5cm,
    xlabel={Input Resolution (pixels)},
    ylabel={GPU Memory (GB)},
    xlabel style={font=\small\bfseries, yshift=-18pt},
    ylabel style={font=\small\bfseries},
    axis line style={thick},
    tick label style={font=\small},
    xmin=400, xmax=4200,
    ymin=0, ymax=22,
    xtick={448,1024,2048,4096},
    xticklabel style={
        rotate=40,
        anchor=north east,
        font=\small,
        /pgf/number format/set thousands separator={}
    },
    grid=major,
    grid style={dashed, gray!40},
    legend style={
        font=\small,
        at={(0.02,0.98)},
        anchor=north west,
        draw=none
    },
    enlarge x limits=0.08
]

\addplot[
    color=caviscolor,
    mark=*,
    mark size=2.8,
    line width=1.8pt
]
coordinates {
    (448,2.41)
    (1024,4.55)
    (2048,8.61)
    (4096,19.44)
};
\addlegendentry{CAVIS}

\addplot[
    color=videomtcolor,
    mark=square*,
    mark size=2.8,
    line width=1.8pt
]
coordinates {
    (448,1.74)
    (1024,2.84)
    (2048,4.92)
    (4096,9.51)
};
\addlegendentry{VidEoMT}

\end{axis}
\end{tikzpicture}
\caption{\textbf{GPU memory consumption of CAVIS and VidEoMT.} Across increasing input resolutions.}
\label{fig:memory_scaling}
\end{figure}
\subsection{Comparison of Memory Consumption}
\cref{fig:memory_scaling} compares the GPU memory consumption of CAVIS and VidEoMT at different input resolutions. As the resolution increases, the memory usage of both methods grows, but VidEoMT consistently requires substantially less memory than CAVIS. The gap becomes more pronounced at higher resolutions. For instance, at a resolution of 4096 pixels, CAVIS consumes 19.44 GB of memory, whereas VidEoMT requires only 9.51 GB. This demonstrates that VidEoMT scales more efficiently with increased spatial resolution.  

\section{Qualitative Results}
\label{sec:supp:qual_results}
\begin{figure*}[t]
\centering

\begin{minipage}[t]{0.23\linewidth}
    \centering
    \includegraphics[width=\linewidth, height=1.5cm]{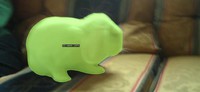}
    \includegraphics[width=\linewidth, height=1.5cm]{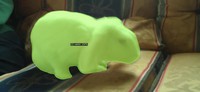}
    \includegraphics[width=\linewidth, height=1.5cm]{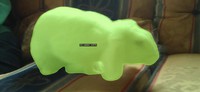}
    \includegraphics[width=\linewidth, height=1.5cm]{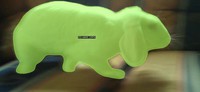}
    \includegraphics[width=\linewidth, height=1.5cm]{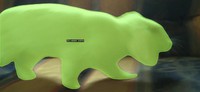}
    \includegraphics[width=\linewidth, height=1.5cm]{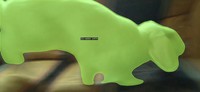}
    \includegraphics[width=\linewidth, height=1.5cm]{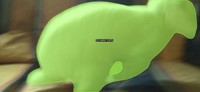}
    \includegraphics[width=\linewidth, height=1.5cm]{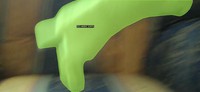}
    \includegraphics[width=\linewidth, height=1.5cm]{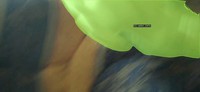}
    \includegraphics[width=\linewidth, height=1.5cm]{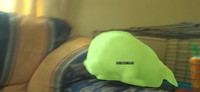}
    \includegraphics[width=\linewidth, height=1.5cm]{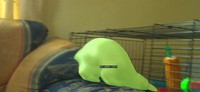}
    \includegraphics[width=\linewidth, height=1.5cm]{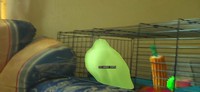}
    \includegraphics[width=\linewidth, height=1.5cm]{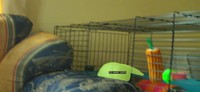}
    \caption*{CAVIS (15 FPS)}
\end{minipage}
\begin{minipage}[t]{0.23\linewidth}
    \centering
      \includegraphics[width=\linewidth, height=1.5cm]{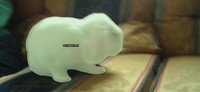}
    \includegraphics[width=\linewidth, height=1.5cm]{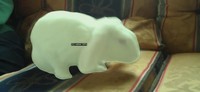}
    \includegraphics[width=\linewidth, height=1.5cm]{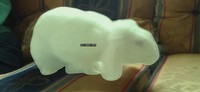}
    \includegraphics[width=\linewidth, height=1.5cm]{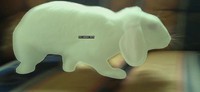}
    \includegraphics[width=\linewidth, height=1.5cm]{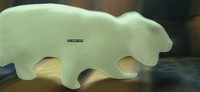}
    \includegraphics[width=\linewidth, height=1.5cm]{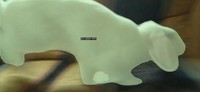}
    \includegraphics[width=\linewidth, height=1.5cm]{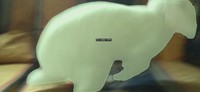}
    \includegraphics[width=\linewidth, height=1.5cm]{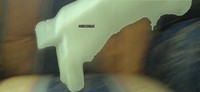}
    \includegraphics[width=\linewidth, height=1.5cm]{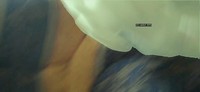}
    \includegraphics[width=\linewidth, height=1.5cm]{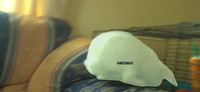}
    \includegraphics[width=\linewidth, height=1.5cm]{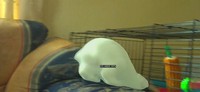}
    \includegraphics[width=\linewidth, height=1.5cm]{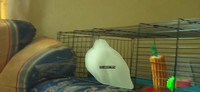}
    \includegraphics[width=\linewidth, height=1.5cm]{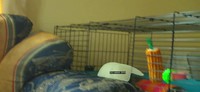}
    \caption*{\textbf{VidEoMT (160 FPS)}}
\end{minipage}
\begin{minipage}[t]{0.23\linewidth}
    \centering
    \includegraphics[width=\linewidth, height=1.5cm]{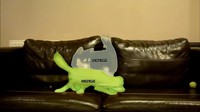}
    \includegraphics[width=\linewidth, height=1.5cm]{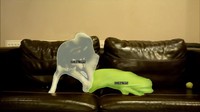}
    \includegraphics[width=\linewidth, height=1.5cm]{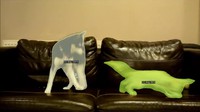}
    \includegraphics[width=\linewidth, height=1.5cm]{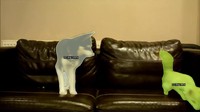}
    \includegraphics[width=\linewidth, height=1.5cm]{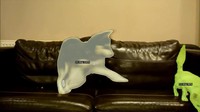}
    \includegraphics[width=\linewidth, height=1.5cm]{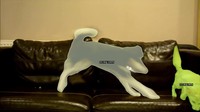}
    \includegraphics[width=\linewidth, height=1.5cm]{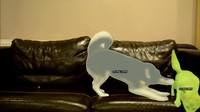}
    \includegraphics[width=\linewidth, height=1.5cm]{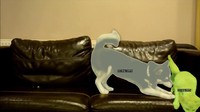}
    \includegraphics[width=\linewidth, height=1.5cm]{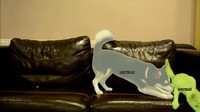}
    \includegraphics[width=\linewidth, height=1.5cm]{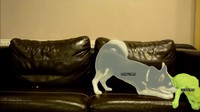}
    \includegraphics[width=\linewidth, height=1.5cm]{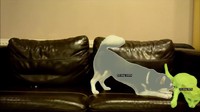}
    \includegraphics[width=\linewidth, height=1.5cm]{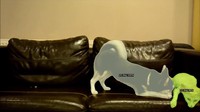}
    \includegraphics[width=\linewidth, height=1.5cm]{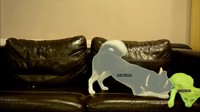}
    \caption*{CAVIS (15 FPS)}
\end{minipage}
\begin{minipage}[t]{0.23\linewidth}
    \centering
    \includegraphics[width=\linewidth, height=1.5cm]{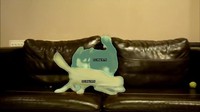}
    \includegraphics[width=\linewidth, height=1.5cm]{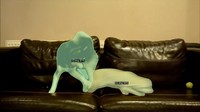}
    \includegraphics[width=\linewidth, height=1.5cm]{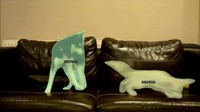}
    \includegraphics[width=\linewidth, height=1.5cm]{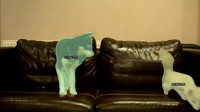}
    \includegraphics[width=\linewidth, height=1.5cm]{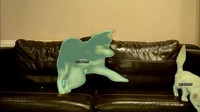}
    \includegraphics[width=\linewidth, height=1.5cm]{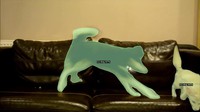}
    \includegraphics[width=\linewidth, height=1.5cm]{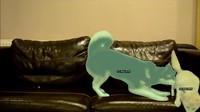}
    \includegraphics[width=\linewidth, height=1.5cm]{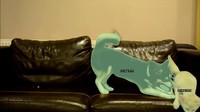}
    \includegraphics[width=\linewidth, height=1.5cm]{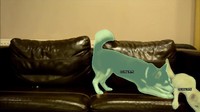}
    \includegraphics[width=\linewidth, height=1.5cm]{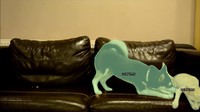}
    \includegraphics[width=\linewidth, height=1.5cm]{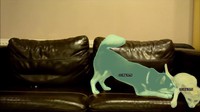}
    \includegraphics[width=\linewidth, height=1.5cm]{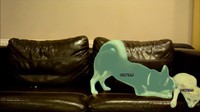}
    \includegraphics[width=\linewidth, height=1.5cm]{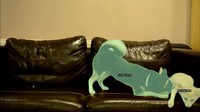}
    \caption*{\textbf{VidEoMT (160 FPS)}}
\end{minipage}
\caption{\textbf{Qualitative results for video instance segmentation.} We compare CAVIS~\citep{lee2025cavis} to VidEoMT on the YouTube-VIS 2019 dataset~\citep{yang2019video}.}
\label{app:fig:qual_yt_2019}
\end{figure*}

\begin{figure*}[t]
\centering

\begin{minipage}[t]{0.23\linewidth}
    \centering
    \includegraphics[width=\linewidth, height=1.5cm]{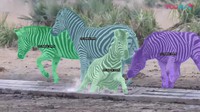}
    \includegraphics[width=\linewidth, height=1.5cm]{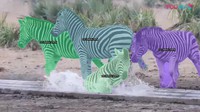}
    \includegraphics[width=\linewidth, height=1.5cm]{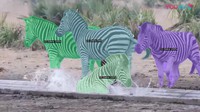}
    \includegraphics[width=\linewidth, height=1.5cm]
    {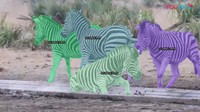}
    \includegraphics[width=\linewidth, height=1.5cm]{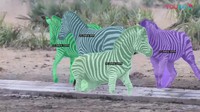}
    \includegraphics[width=\linewidth, height=1.5cm]{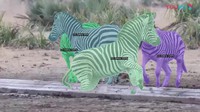}
    \includegraphics[width=\linewidth, height=1.5cm]{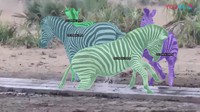}
    \includegraphics[width=\linewidth, height=1.5cm]{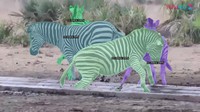}
    \includegraphics[width=\linewidth, height=1.5cm]{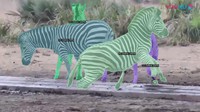}
    \includegraphics[width=\linewidth, height=1.5cm]{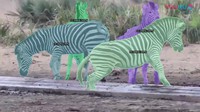}
    \includegraphics[width=\linewidth, height=1.5cm]{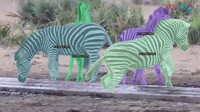}
    \includegraphics[width=\linewidth, height=1.5cm]{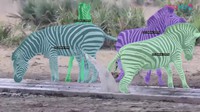}
    \includegraphics[width=\linewidth, height=1.5cm]{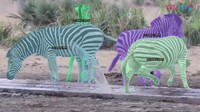}
    \caption*{CAVIS (15 FPS)}
\end{minipage}
\begin{minipage}[t]{0.23\linewidth}
    \centering
\includegraphics[width=\linewidth, height=1.5cm]{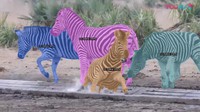}
    \includegraphics[width=\linewidth, height=1.5cm]{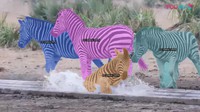}
    \includegraphics[width=\linewidth, height=1.5cm]{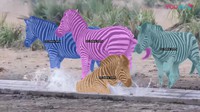}
    \includegraphics[width=\linewidth, height=1.5cm]
    {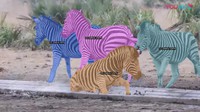}
    \includegraphics[width=\linewidth, height=1.5cm]{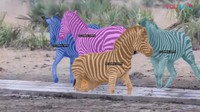}
    \includegraphics[width=\linewidth, height=1.5cm]{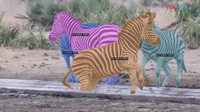}
    \includegraphics[width=\linewidth, height=1.5cm]{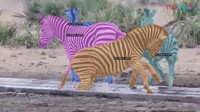}
    \includegraphics[width=\linewidth, height=1.5cm]{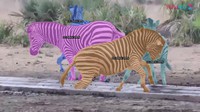}
    \includegraphics[width=\linewidth, height=1.5cm]{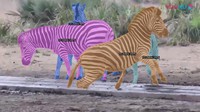}
    \includegraphics[width=\linewidth, height=1.5cm]{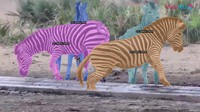}
    \includegraphics[width=\linewidth, height=1.5cm]{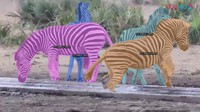}
    \includegraphics[width=\linewidth, height=1.5cm]{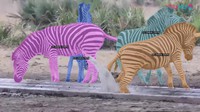}
    \includegraphics[width=\linewidth, height=1.5cm]{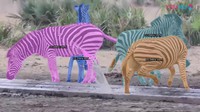}
    \caption*{\textbf{VidEoMT (112 FPS)}}
\end{minipage}
\begin{minipage}[t]{0.23\linewidth}
    \centering
    \includegraphics[width=\linewidth, height=1.5cm]{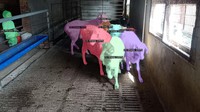}
    \includegraphics[width=\linewidth, height=1.5cm]{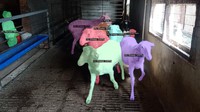}
    \includegraphics[width=\linewidth, height=1.5cm]{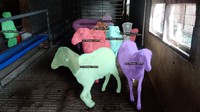}
    \includegraphics[width=\linewidth, height=1.5cm]{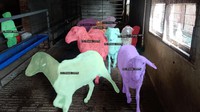}
    \includegraphics[width=\linewidth, height=1.5cm]{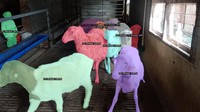}
    \includegraphics[width=\linewidth, height=1.5cm]{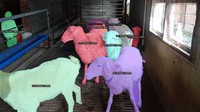}
    \includegraphics[width=\linewidth, height=1.5cm]{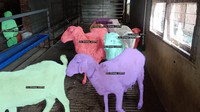}
    \includegraphics[width=\linewidth, height=1.5cm]{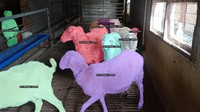}
    \includegraphics[width=\linewidth, height=1.5cm]{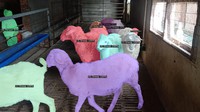}
    \includegraphics[width=\linewidth, height=1.5cm]{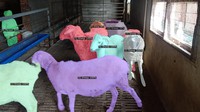}
    \includegraphics[width=\linewidth, height=1.5cm]{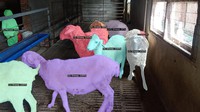}
    \includegraphics[width=\linewidth, height=1.5cm]{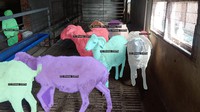}
    \includegraphics[width=\linewidth, height=1.5cm]{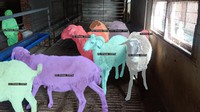}
    \caption*{CAVIS (15 FPS)}
\end{minipage}
\begin{minipage}[t]{0.23\linewidth}
    \centering
    \includegraphics[width=\linewidth, height=1.5cm]{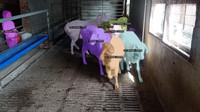}
    \includegraphics[width=\linewidth, height=1.5cm]{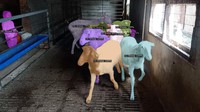}
    \includegraphics[width=\linewidth, height=1.5cm]{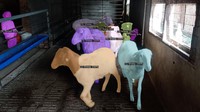}
    \includegraphics[width=\linewidth, height=1.5cm]{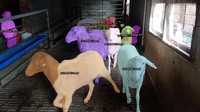}
    \includegraphics[width=\linewidth, height=1.5cm]{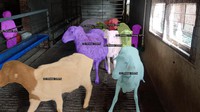}
    \includegraphics[width=\linewidth, height=1.5cm]{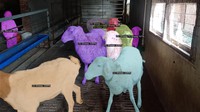}
    \includegraphics[width=\linewidth, height=1.5cm]{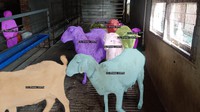}
    \includegraphics[width=\linewidth, height=1.5cm]{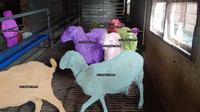}
    \includegraphics[width=\linewidth, height=1.5cm]{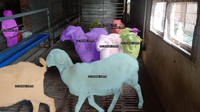}
    \includegraphics[width=\linewidth, height=1.5cm]{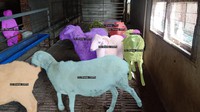}
    \includegraphics[width=\linewidth, height=1.5cm]{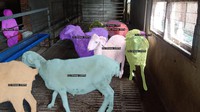}
    \includegraphics[width=\linewidth, height=1.5cm]{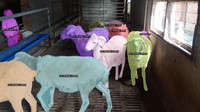}
    \includegraphics[width=\linewidth, height=1.5cm]{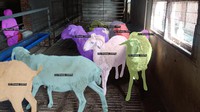}
    \caption*{\textbf{VidEoMT (112 FPS)}}
\end{minipage}
\caption{\textbf{Qualitative results for video instance segmentation.} We compare CAVIS~\citep{lee2025cavis} to VidEoMT on the OVIS dataset~\citep{qi2022occluded}.}
\label{app:fig:qual_ovis}
\end{figure*}

\begin{figure*}[t]
\centering

\begin{minipage}[t]{0.23\linewidth}
    \centering
      \includegraphics[width=\linewidth, height=1.5cm]{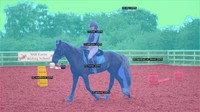}
      \includegraphics[width=\linewidth, height=1.5cm]{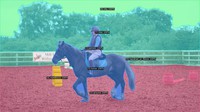}
      \includegraphics[width=\linewidth, height=1.5cm]{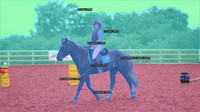}
      \includegraphics[width=\linewidth, height=1.5cm]{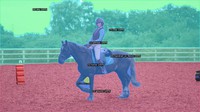}
      \includegraphics[width=\linewidth, height=1.5cm]{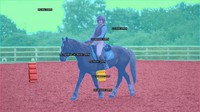}
      \includegraphics[width=\linewidth, height=1.5cm]{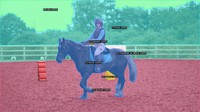}
      \includegraphics[width=\linewidth, height=1.5cm]{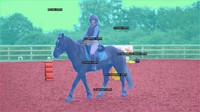}
      \includegraphics[width=\linewidth, height=1.5cm]{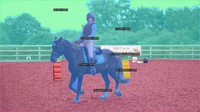}
      \includegraphics[width=\linewidth, height=1.5cm]{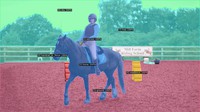}
      \includegraphics[width=\linewidth, height=1.5cm]{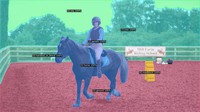}
      \includegraphics[width=\linewidth, height=1.5cm]{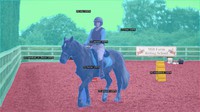}
      \includegraphics[width=\linewidth, height=1.5cm]{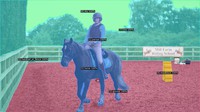}
      \includegraphics[width=\linewidth, height=1.5cm]{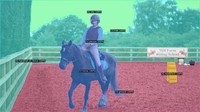}

    \caption*{CAVIS (10 FPS)}
\end{minipage}
\begin{minipage}[t]{0.23\linewidth}
    \centering
        \includegraphics[width=\linewidth, height=1.5cm]{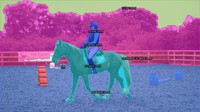}
        \includegraphics[width=\linewidth, height=1.5cm]{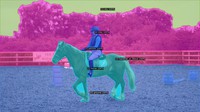}
      \includegraphics[width=\linewidth, height=1.5cm]{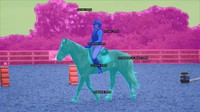}
      \includegraphics[width=\linewidth, height=1.5cm]{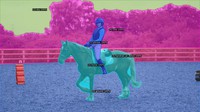}
      \includegraphics[width=\linewidth, height=1.5cm]{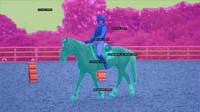}
      \includegraphics[width=\linewidth, height=1.5cm]{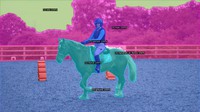}
      \includegraphics[width=\linewidth, height=1.5cm]{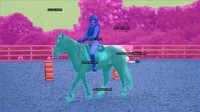}
      \includegraphics[width=\linewidth, height=1.5cm]{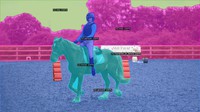}
      \includegraphics[width=\linewidth, height=1.5cm]{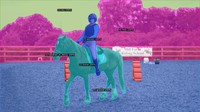}
      \includegraphics[width=\linewidth, height=1.5cm]{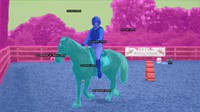}
      \includegraphics[width=\linewidth, height=1.5cm]{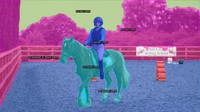}
      \includegraphics[width=\linewidth, height=1.5cm]{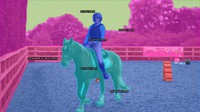}
      \includegraphics[width=\linewidth, height=1.5cm]{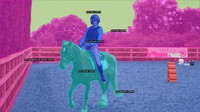}
    
    \caption*{\textbf{VidEoMT (75 FPS)}}
\end{minipage}
\begin{minipage}[t]{0.23\linewidth}
    \centering
     \includegraphics[width=\linewidth, height=1.5cm]{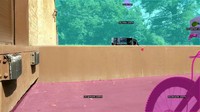}
     \includegraphics[width=\linewidth, height=1.5cm]{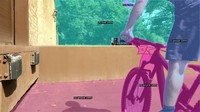}
     \includegraphics[width=\linewidth, height=1.5cm]{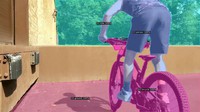}
     \includegraphics[width=\linewidth, height=1.5cm]{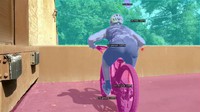}
     \includegraphics[width=\linewidth, height=1.5cm]{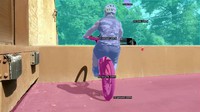}
     \includegraphics[width=\linewidth, height=1.5cm]{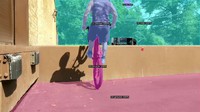}
     \includegraphics[width=\linewidth, height=1.5cm]{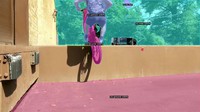}
     \includegraphics[width=\linewidth, height=1.5cm]{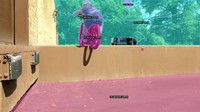}
     \includegraphics[width=\linewidth, height=1.5cm]{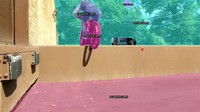}
     \includegraphics[width=\linewidth, height=1.5cm]{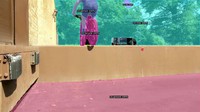}
     \includegraphics[width=\linewidth, height=1.5cm]{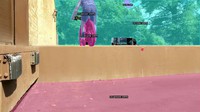}
     \includegraphics[width=\linewidth, height=1.5cm]{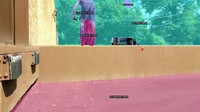}
     \includegraphics[width=\linewidth, height=1.5cm]{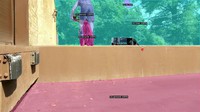}
  
    \caption*{CAVIS (10 FPS)}
\end{minipage}
\begin{minipage}[t]{0.23\linewidth}
    \centering
   \includegraphics[width=\linewidth, height=1.5cm]{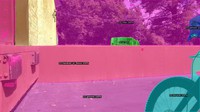}
     \includegraphics[width=\linewidth, height=1.5cm]{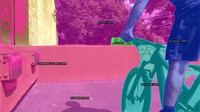}
     \includegraphics[width=\linewidth, height=1.5cm]{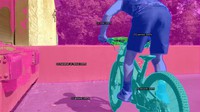}
     \includegraphics[width=\linewidth, height=1.5cm]{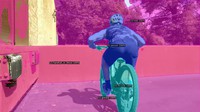}
     \includegraphics[width=\linewidth, height=1.5cm]{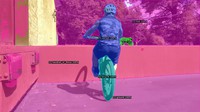}
     \includegraphics[width=\linewidth, height=1.5cm]{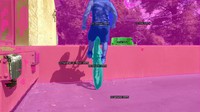}
     \includegraphics[width=\linewidth, height=1.5cm]{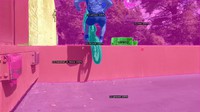}
     \includegraphics[width=\linewidth, height=1.5cm]{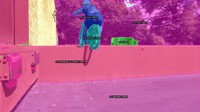}
     \includegraphics[width=\linewidth, height=1.5cm]{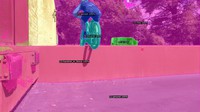}
     \includegraphics[width=\linewidth, height=1.5cm]{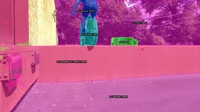}
     \includegraphics[width=\linewidth, height=1.5cm]{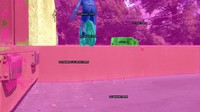}
     \includegraphics[width=\linewidth, height=1.5cm]{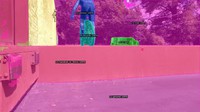}
     \includegraphics[width=\linewidth, height=1.5cm]{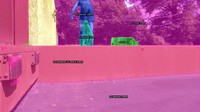}
  
    \caption*{\textbf{VidEoMT (75 FPS)}}
\end{minipage}
\caption{\textbf{Qualitative results for video panoptic segmentation.} We compare CAVIS~\citep{lee2025cavis} to VidEoMT on the VIPSeg dataset~\citep{miao2022large}.}
\label{app:fig:qual_vipseg}
\end{figure*}

In \cref{app:fig:qual_yt_2019,app:fig:qual_ovis,app:fig:qual_vipseg}, we visualize the predictions of both CAVIS~\citep{lee2025cavis} and VidEoMT for VIS and VPS on the YouTube-VIS 2019~\citep{yang2019video}, OVIS~\citep{qi2022occluded}, and VIPSeg~\citep{miao2022large} datasets.

\section{Failure Cases} 
\label{sec:supp:failure_analysis}
\begin{figure*}[t]
\centering
\begin{minipage}[t]{0.45\linewidth}
    \centering
    \includegraphics[width=\linewidth]{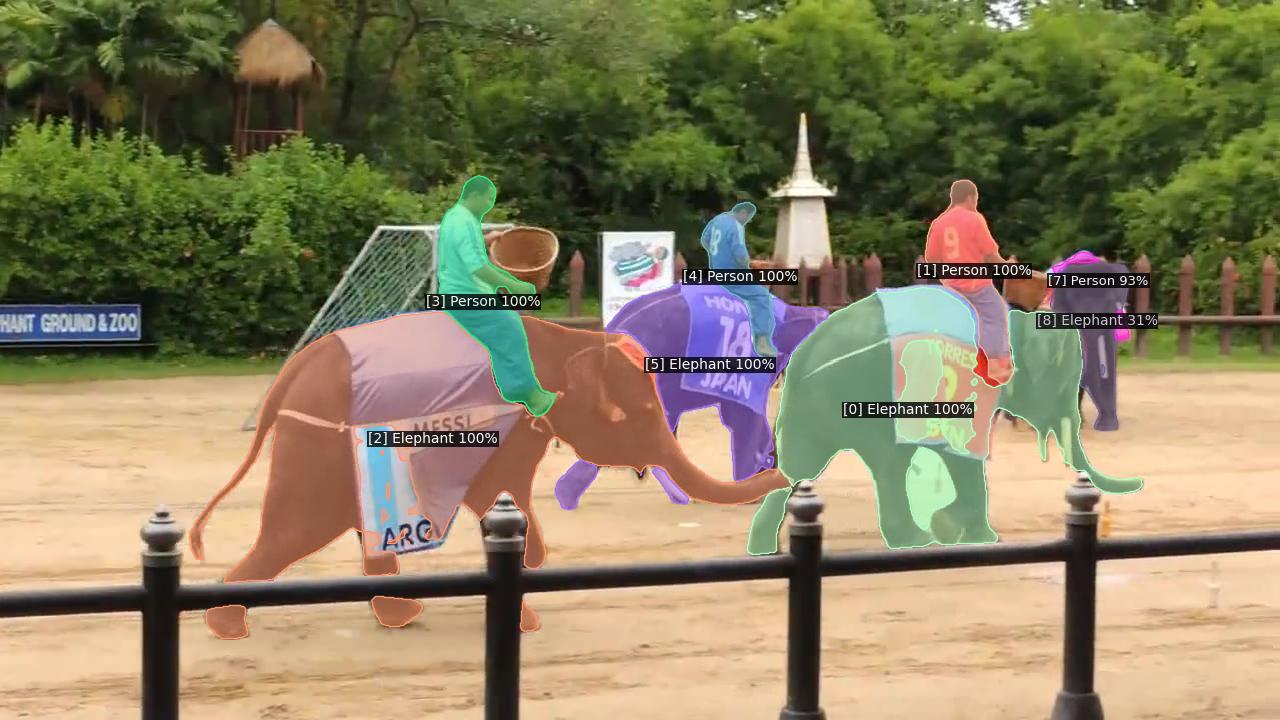}
    \includegraphics[width=\linewidth]{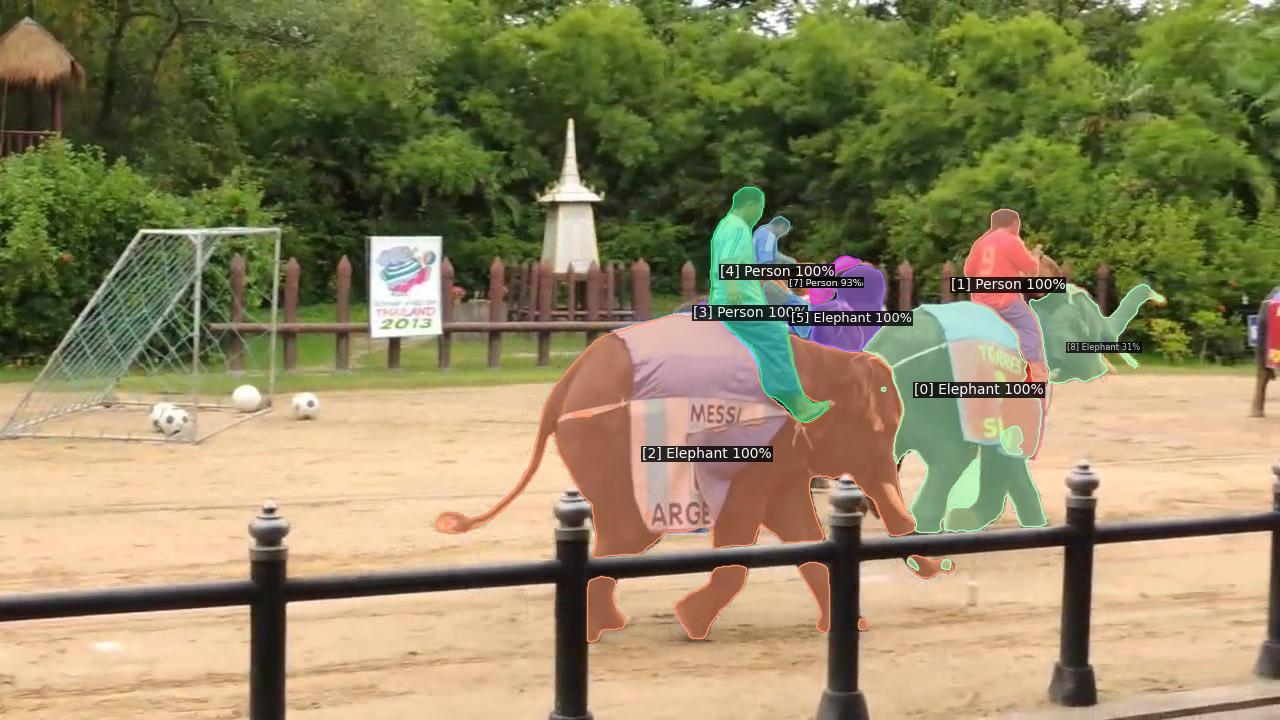}
    \includegraphics[width=\linewidth]{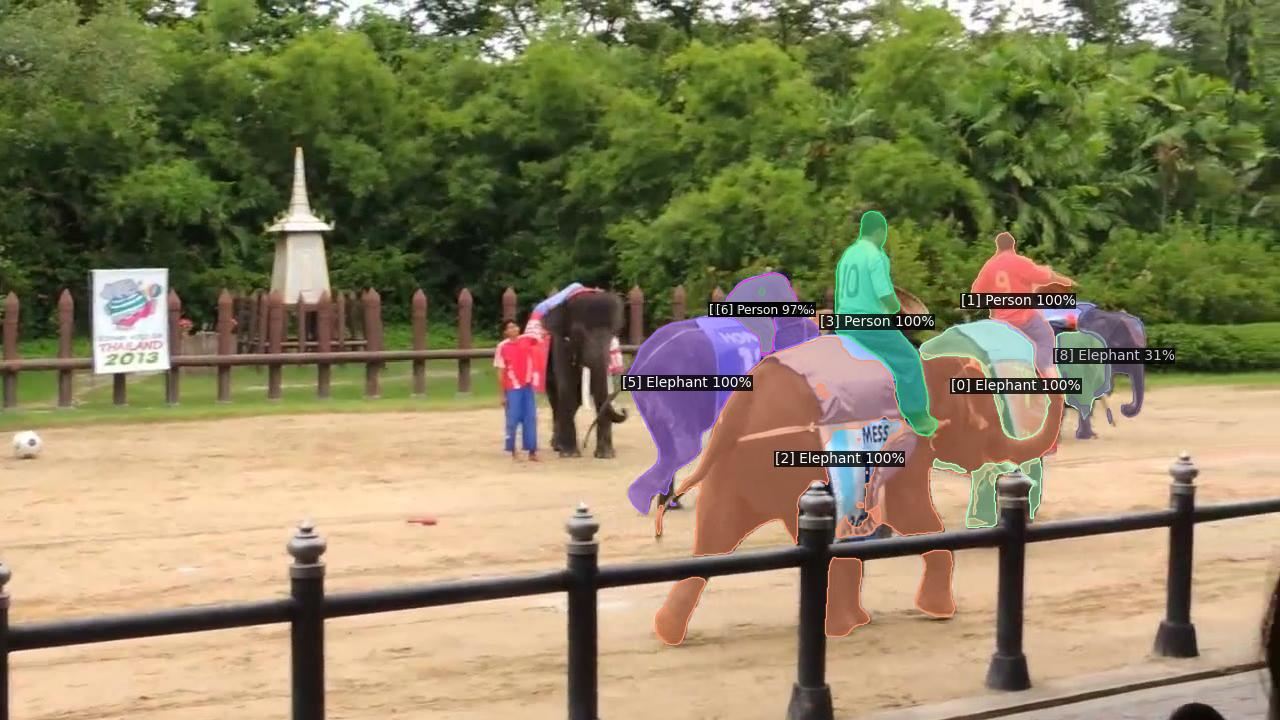}
    \includegraphics[width=\linewidth]{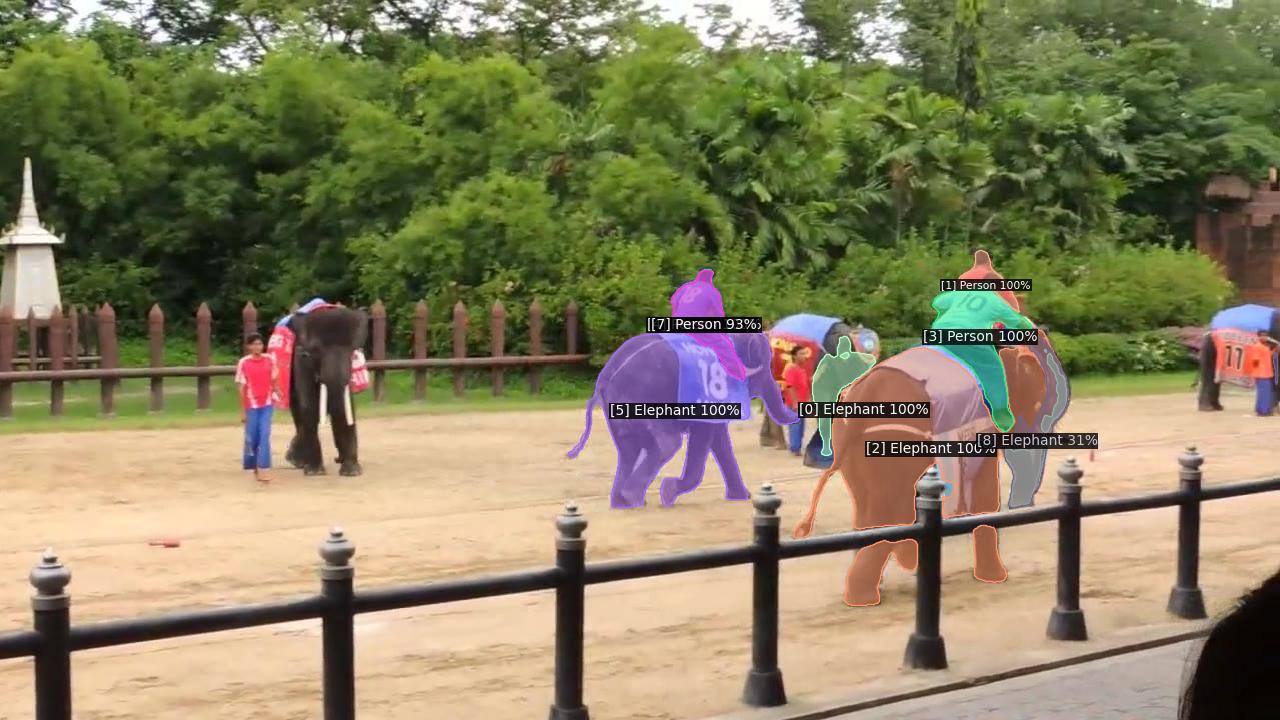}
    \includegraphics[width=\linewidth]{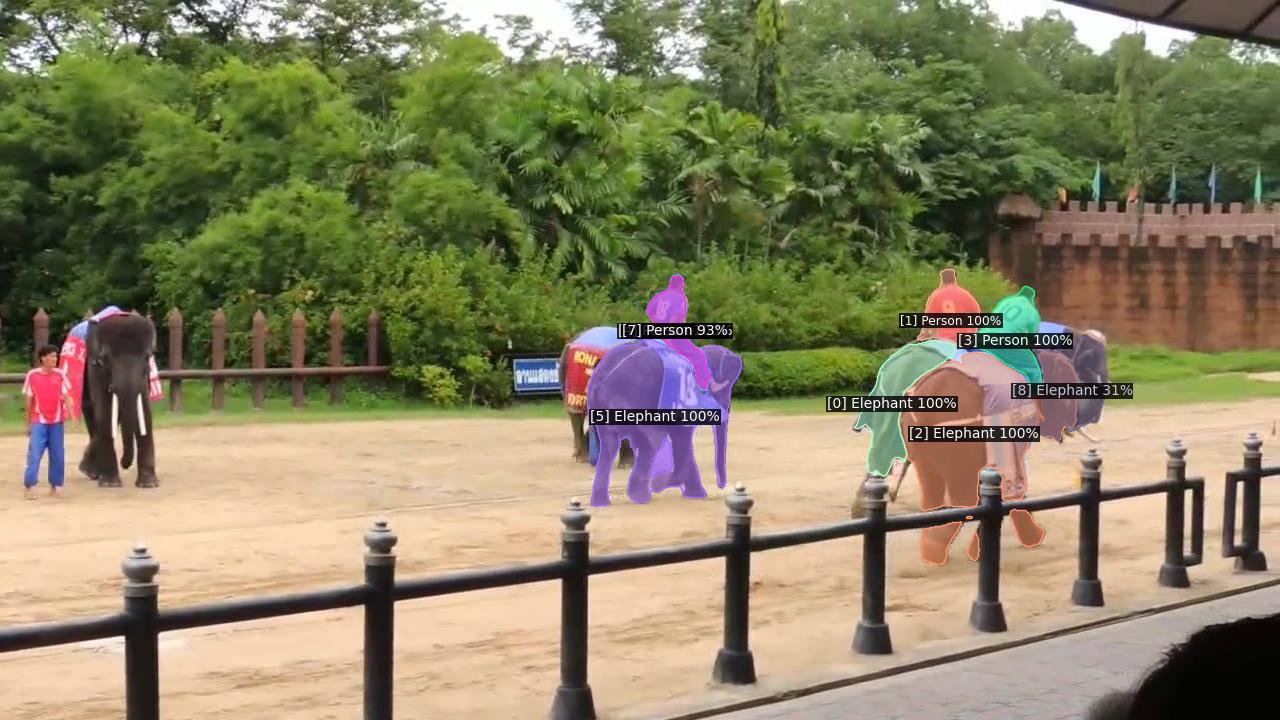}
    \caption*{CAVIS (15 FPS)}
\end{minipage}
\begin{minipage}[t]{0.45\linewidth}
    \centering
    \includegraphics[width=\linewidth]{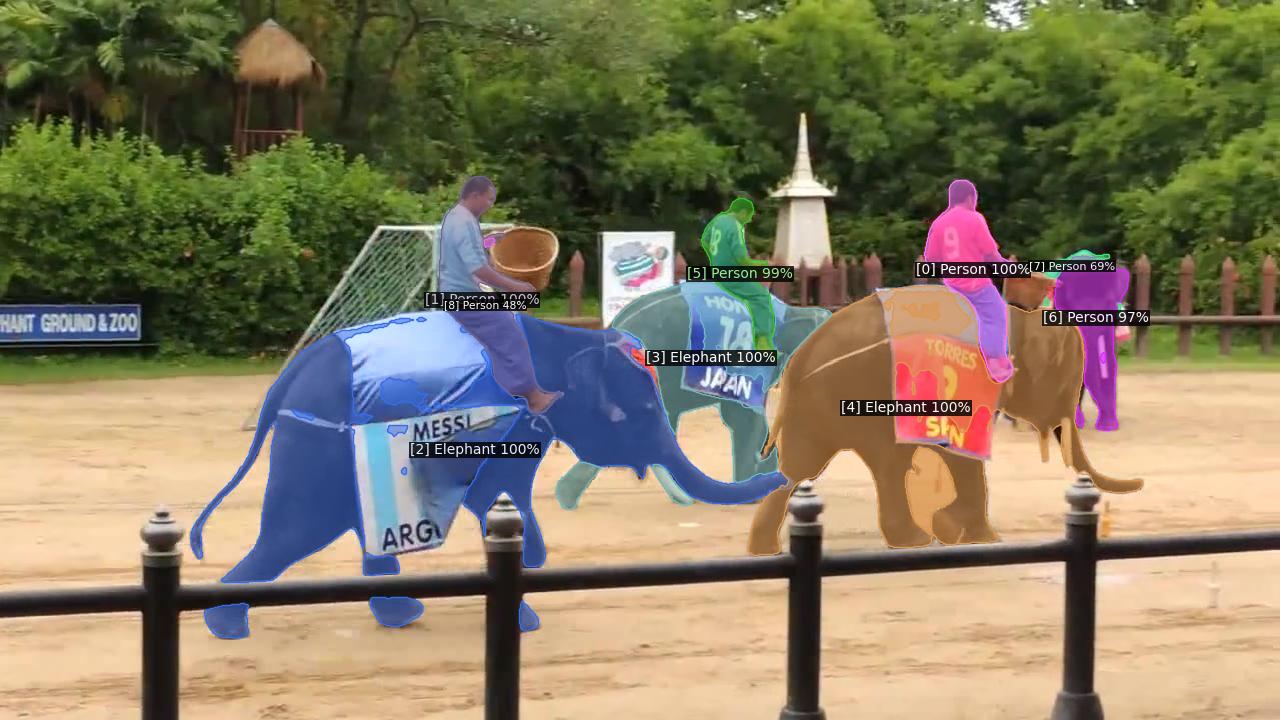}
    \includegraphics[width=\linewidth]{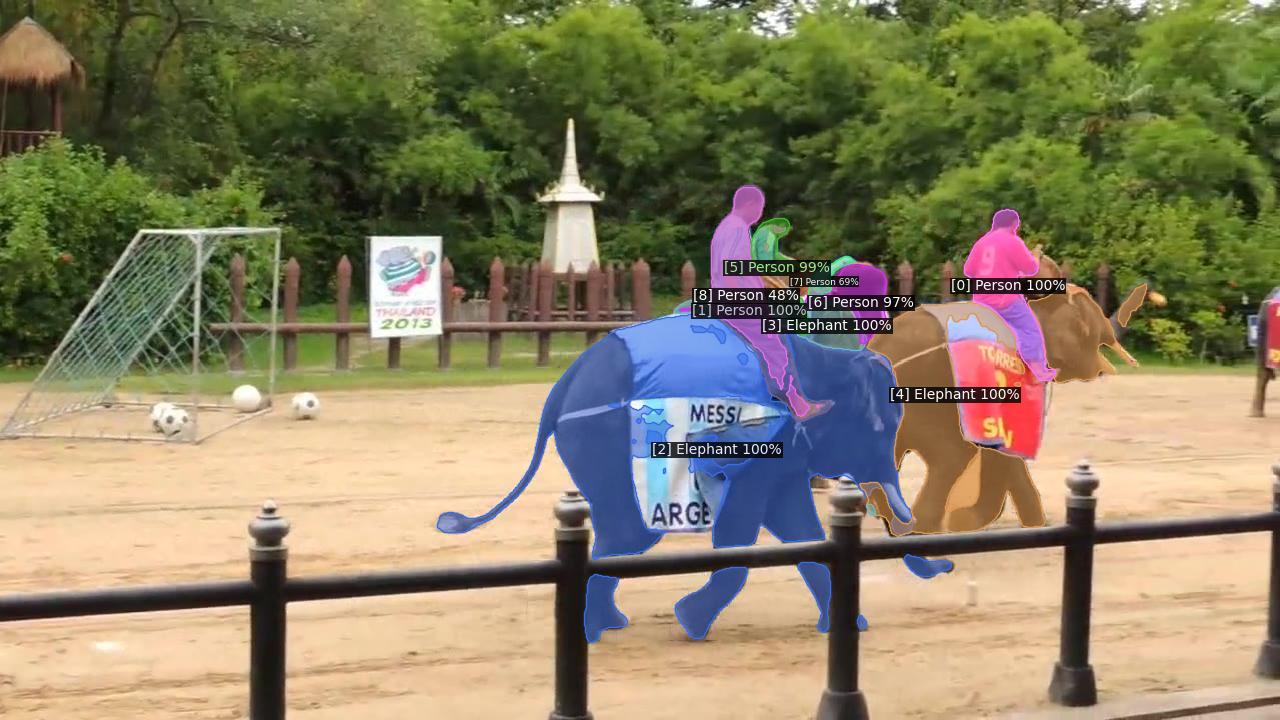}
    \includegraphics[width=\linewidth]{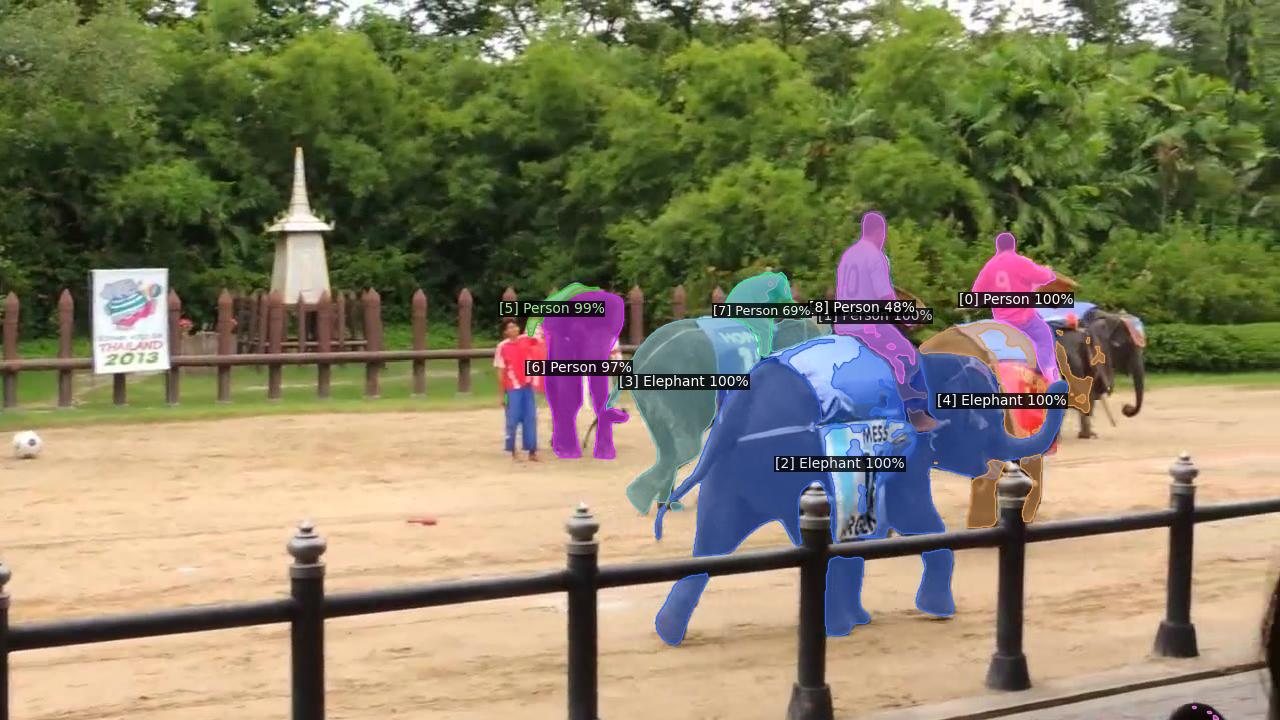}
    \includegraphics[width=\linewidth]{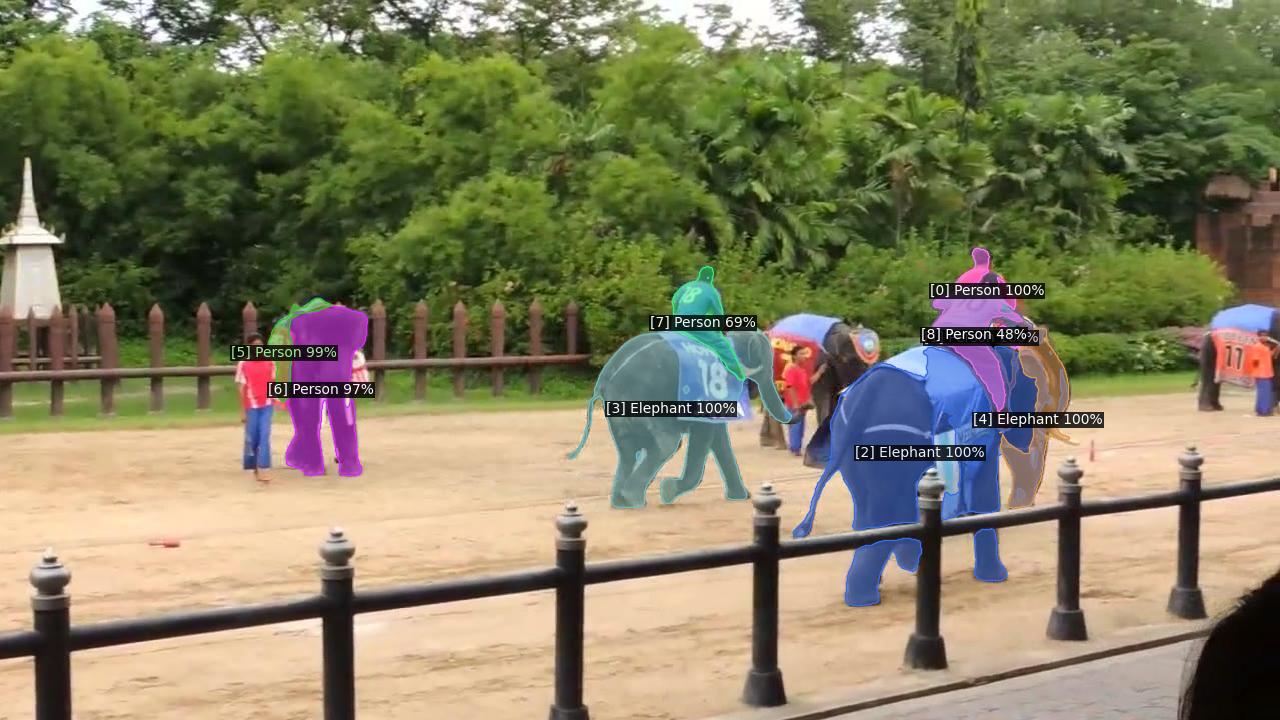}
    \includegraphics[width=\linewidth]{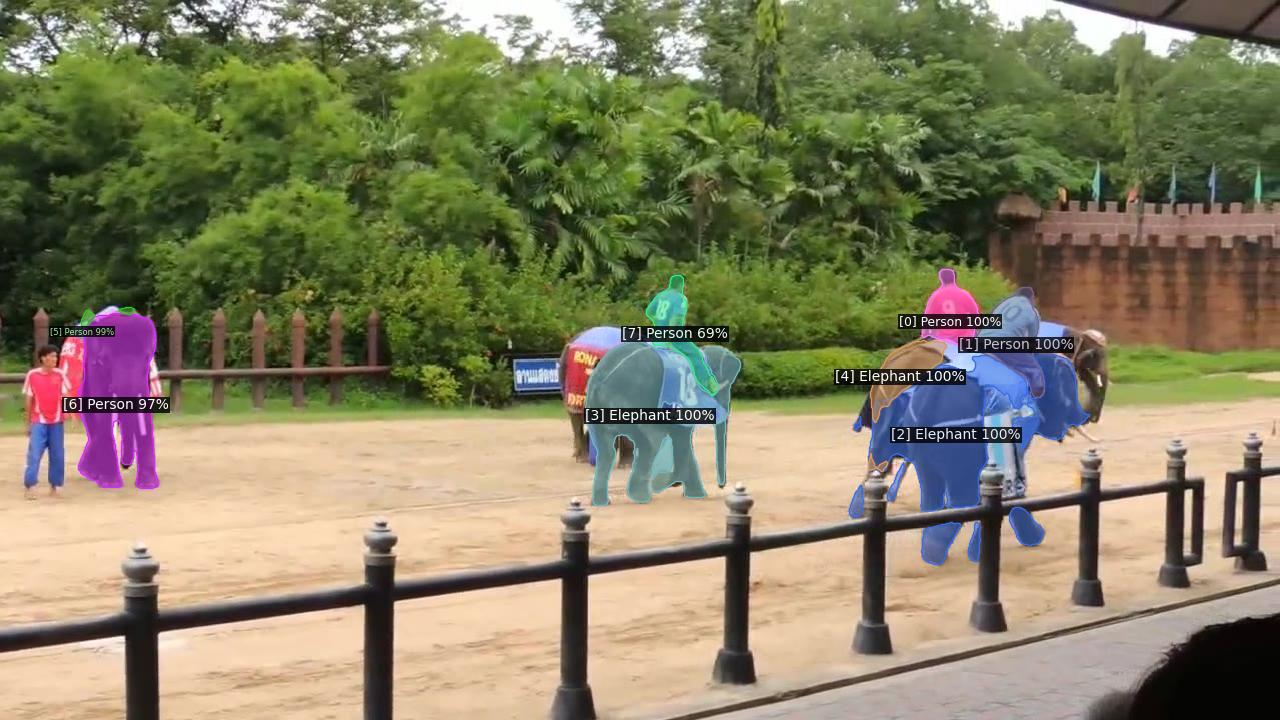}
    \caption*{\textbf{VidEoMT (112 FPS)}}
\end{minipage}
\caption{\textbf{Failure cases.} We illustrate failure cases of CAVIS~\citep{lee2025cavis} and VidEoMT on the OVIS dataset~\citep{qi2022occluded}. In row 3, in a highly occluded scene, identity switching occurs between persons for both models. In row 4, both methods fail to detect newly appearing elephants and persons.}
\label{app:fig:failure_case_1}
\end{figure*}

\begin{figure*}[t]
\centering
\begin{minipage}[t]{0.45\linewidth}
    \centering
    \includegraphics[width=\linewidth]{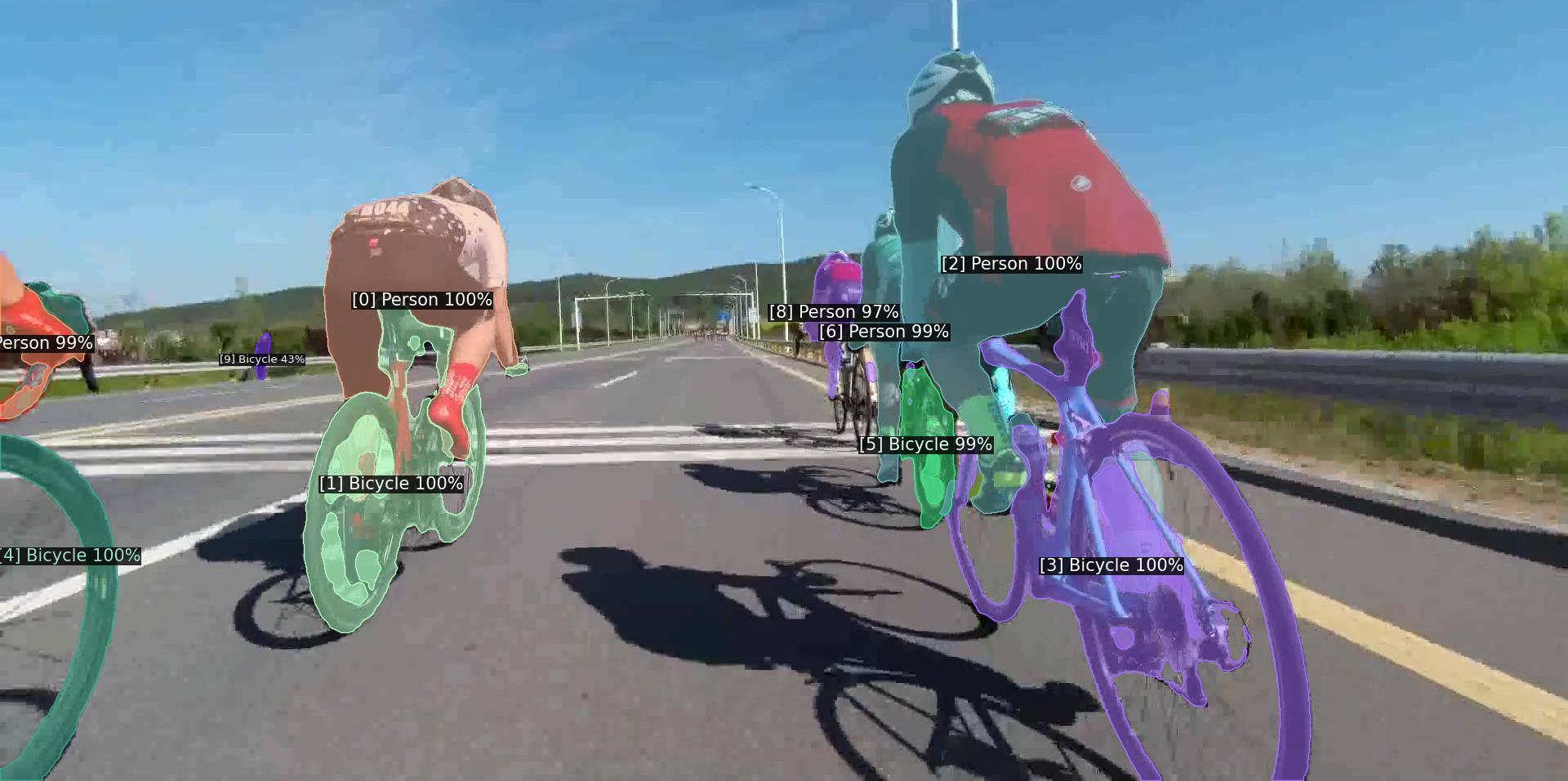}
    \includegraphics[width=\linewidth]{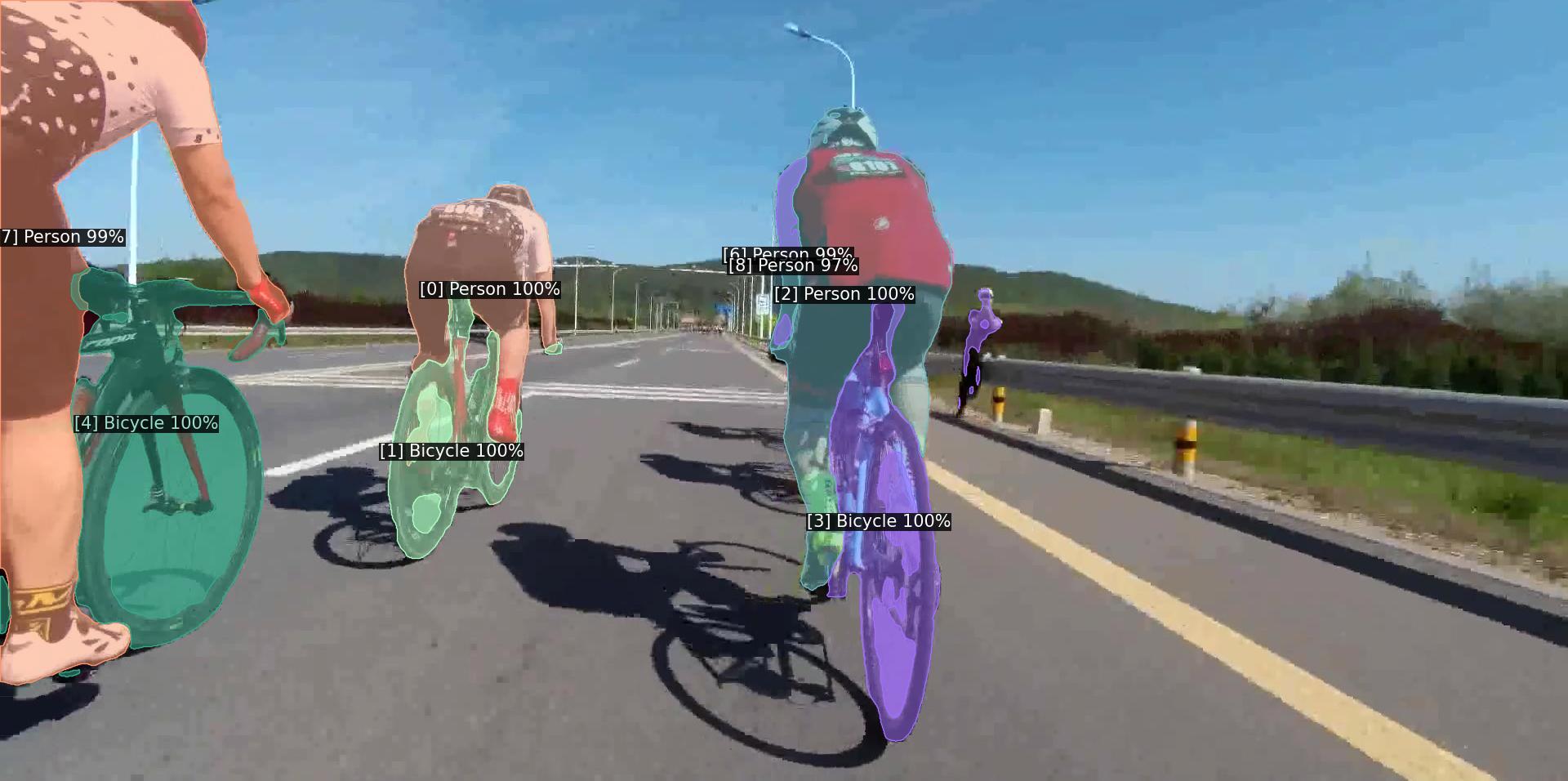}
    \includegraphics[width=\linewidth]{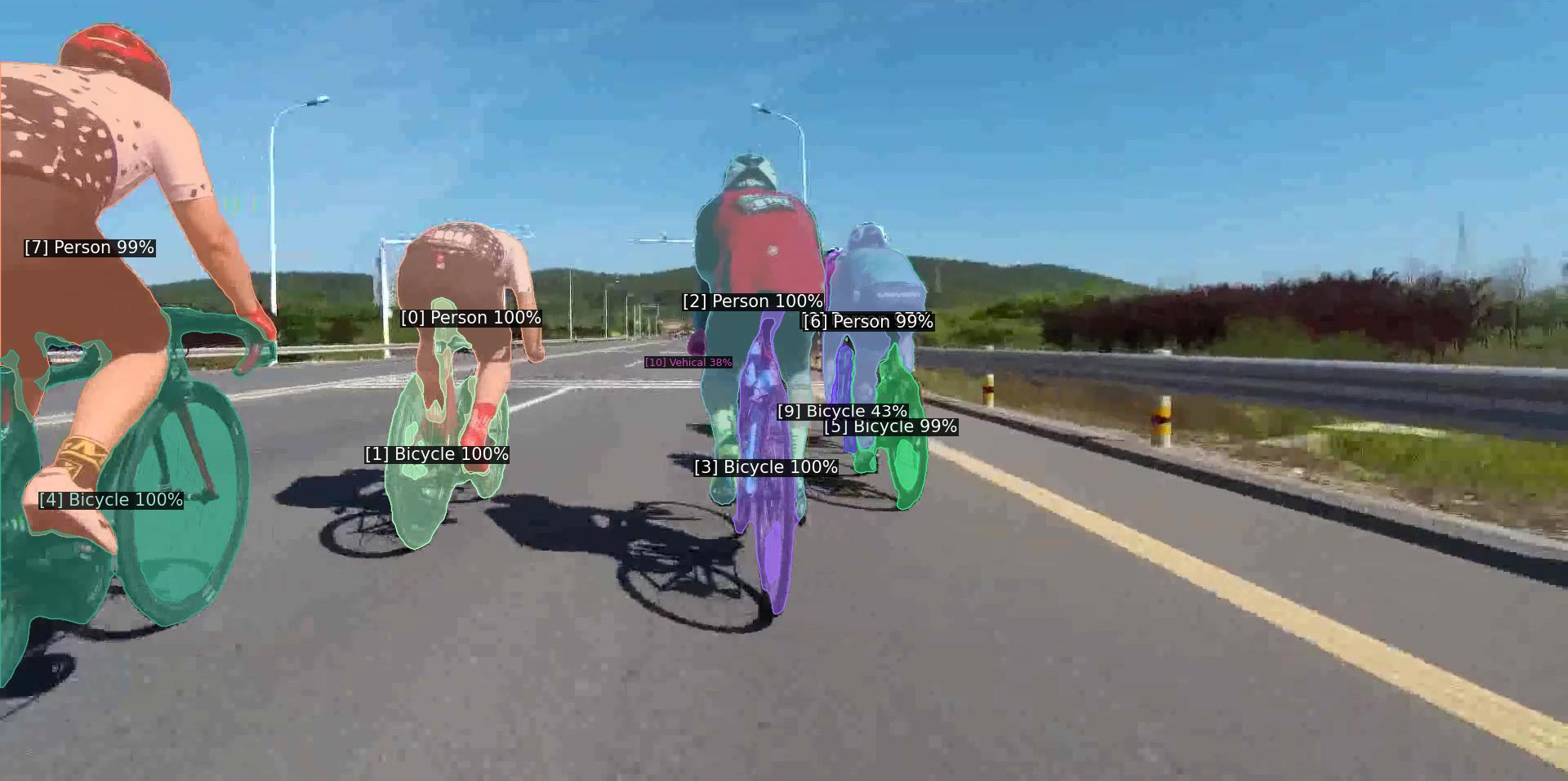}
    \includegraphics[width=\linewidth]{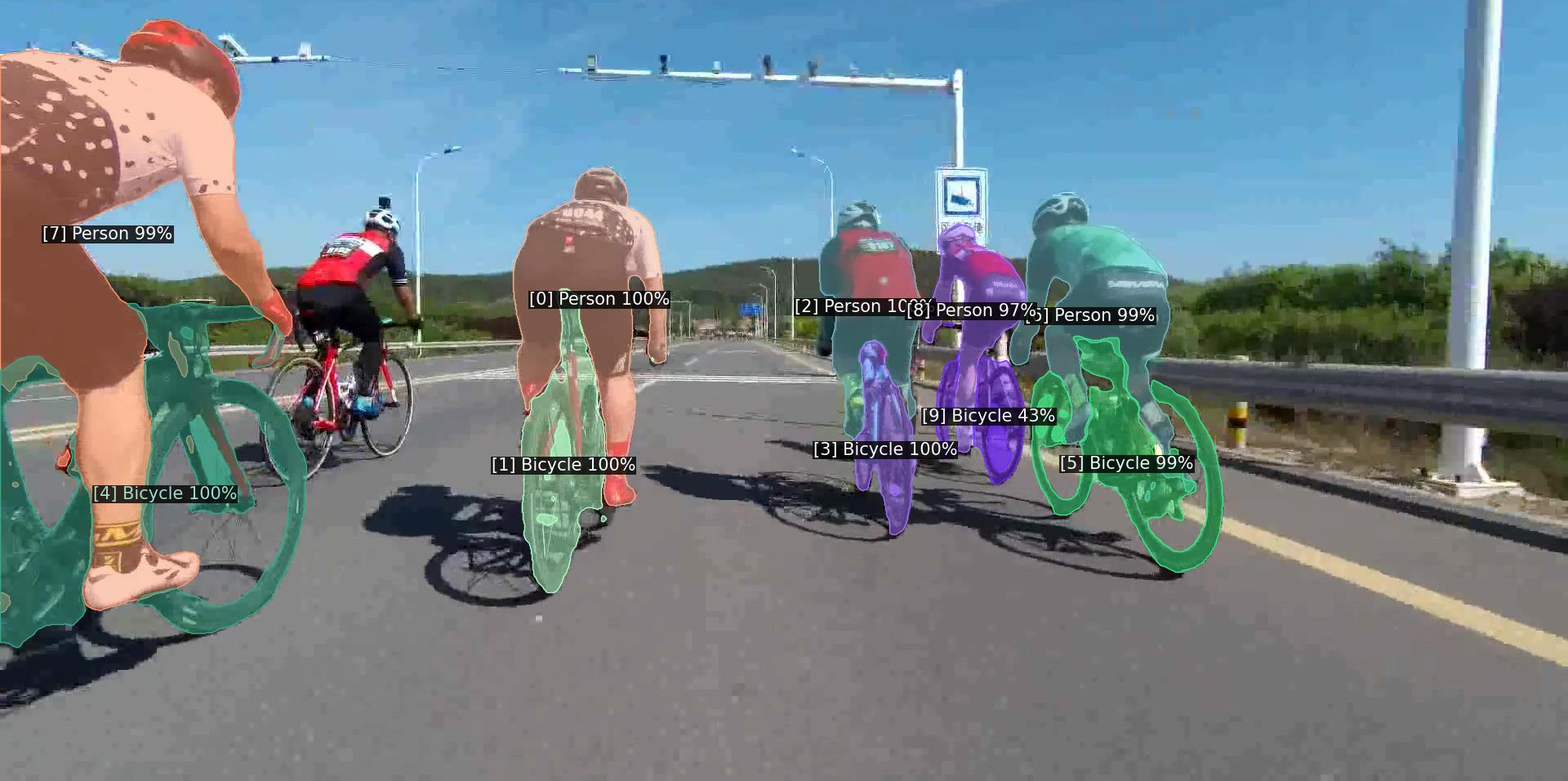}
    \includegraphics[width=\linewidth]{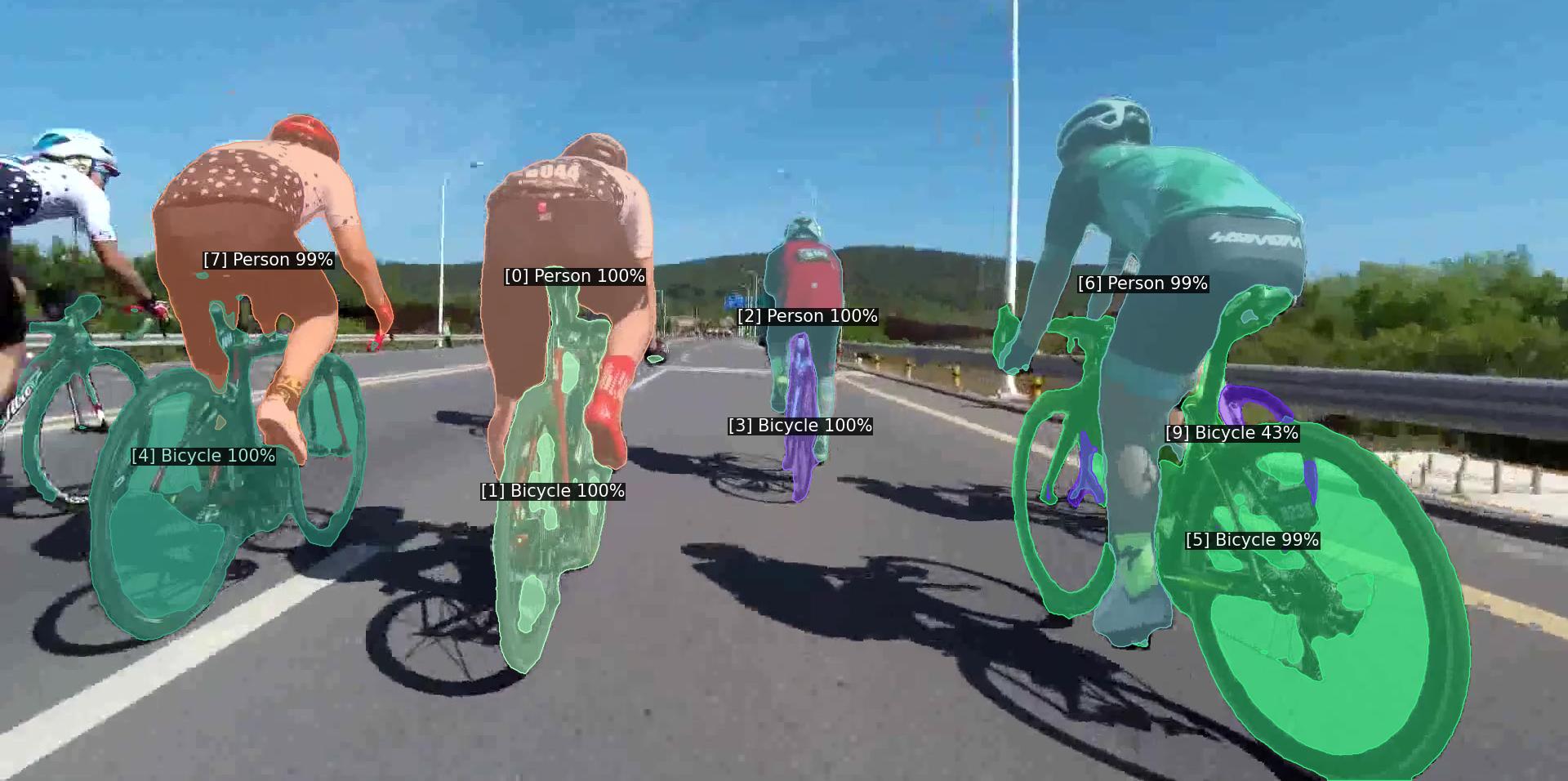}
    \includegraphics[width=\linewidth]{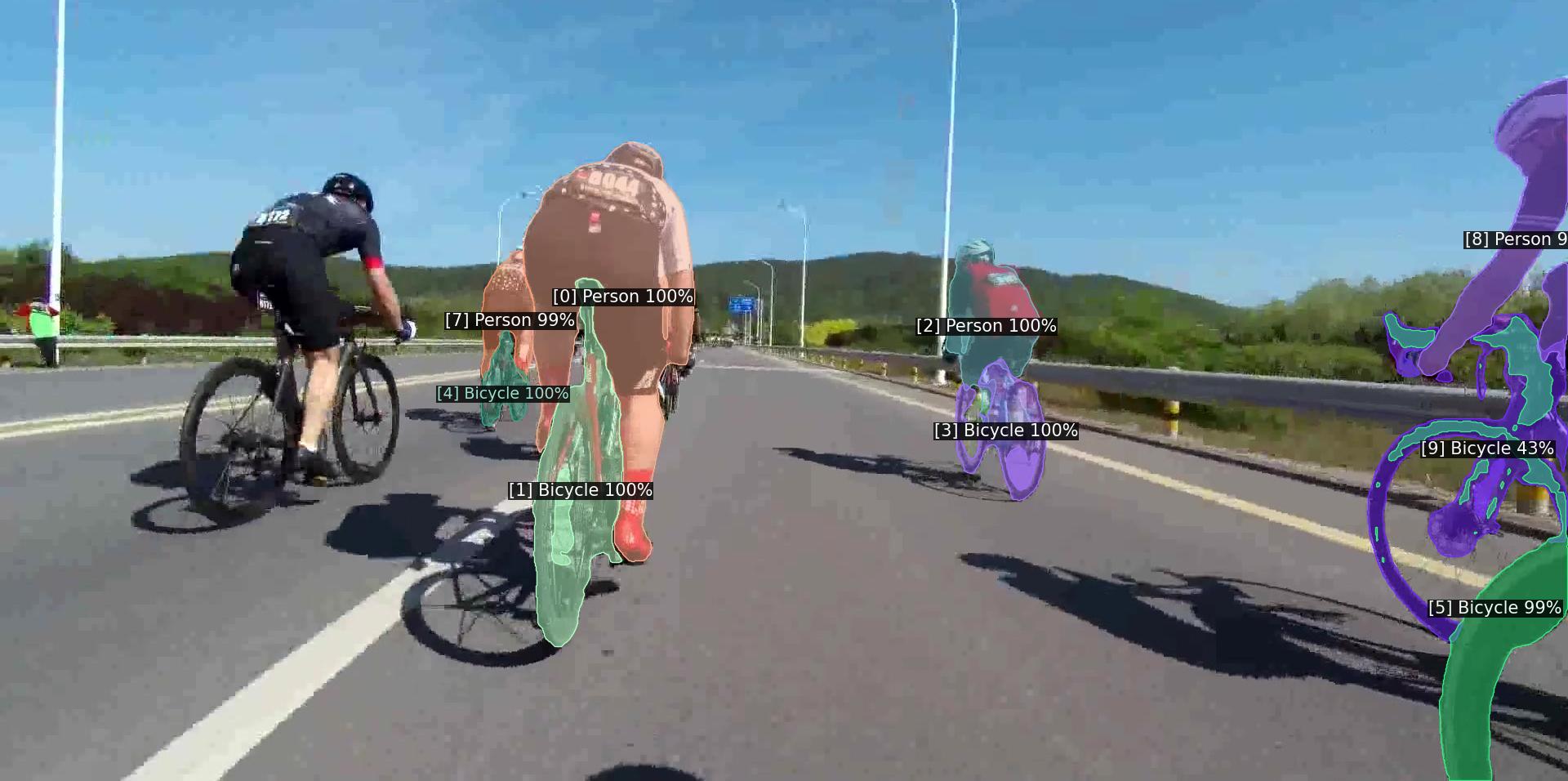}
    \caption*{CAVIS (15 FPS)}
\end{minipage}
\begin{minipage}[t]{0.45\linewidth}
    \centering
    \includegraphics[width=\linewidth]{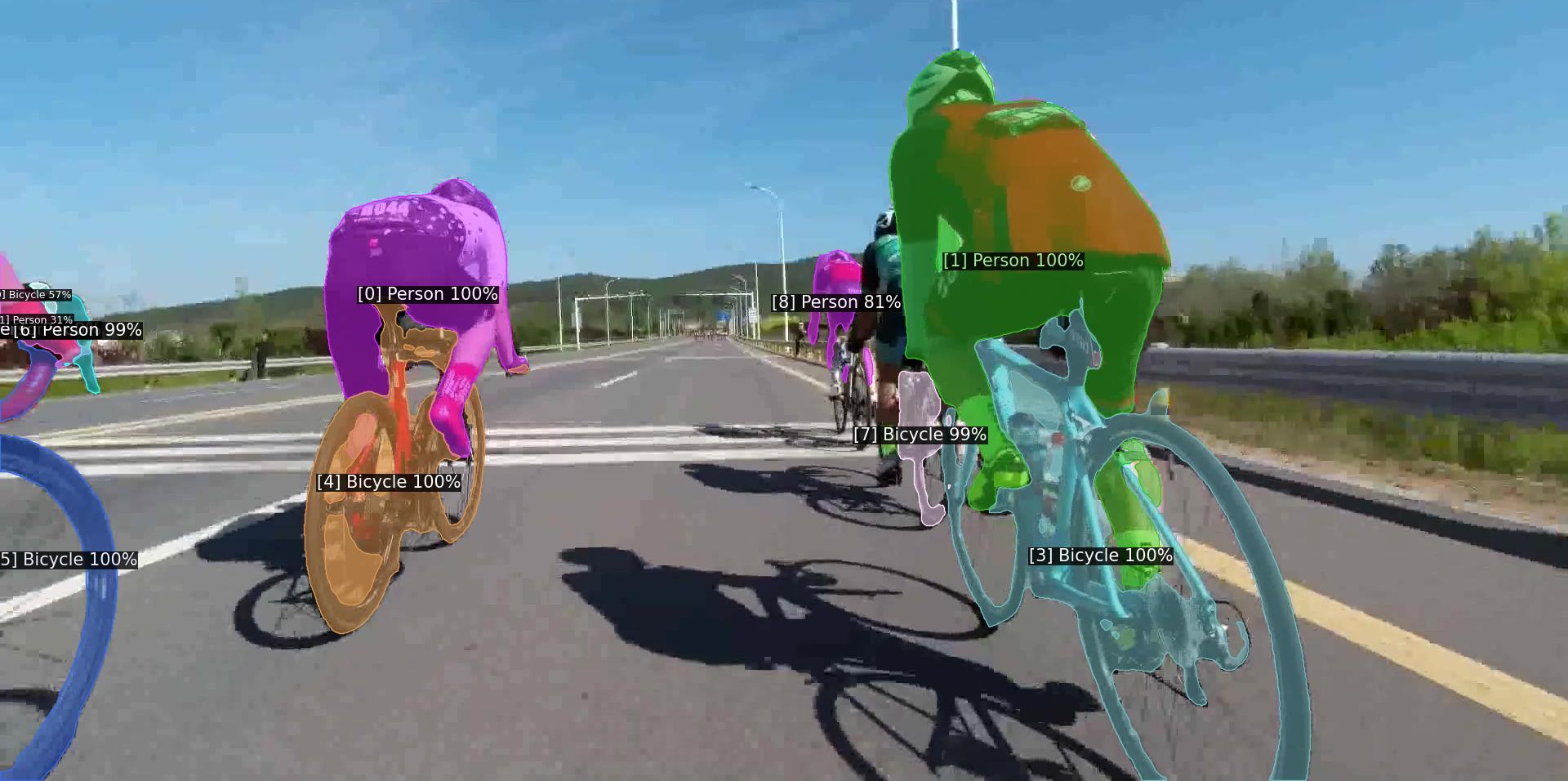}
    \includegraphics[width=\linewidth]{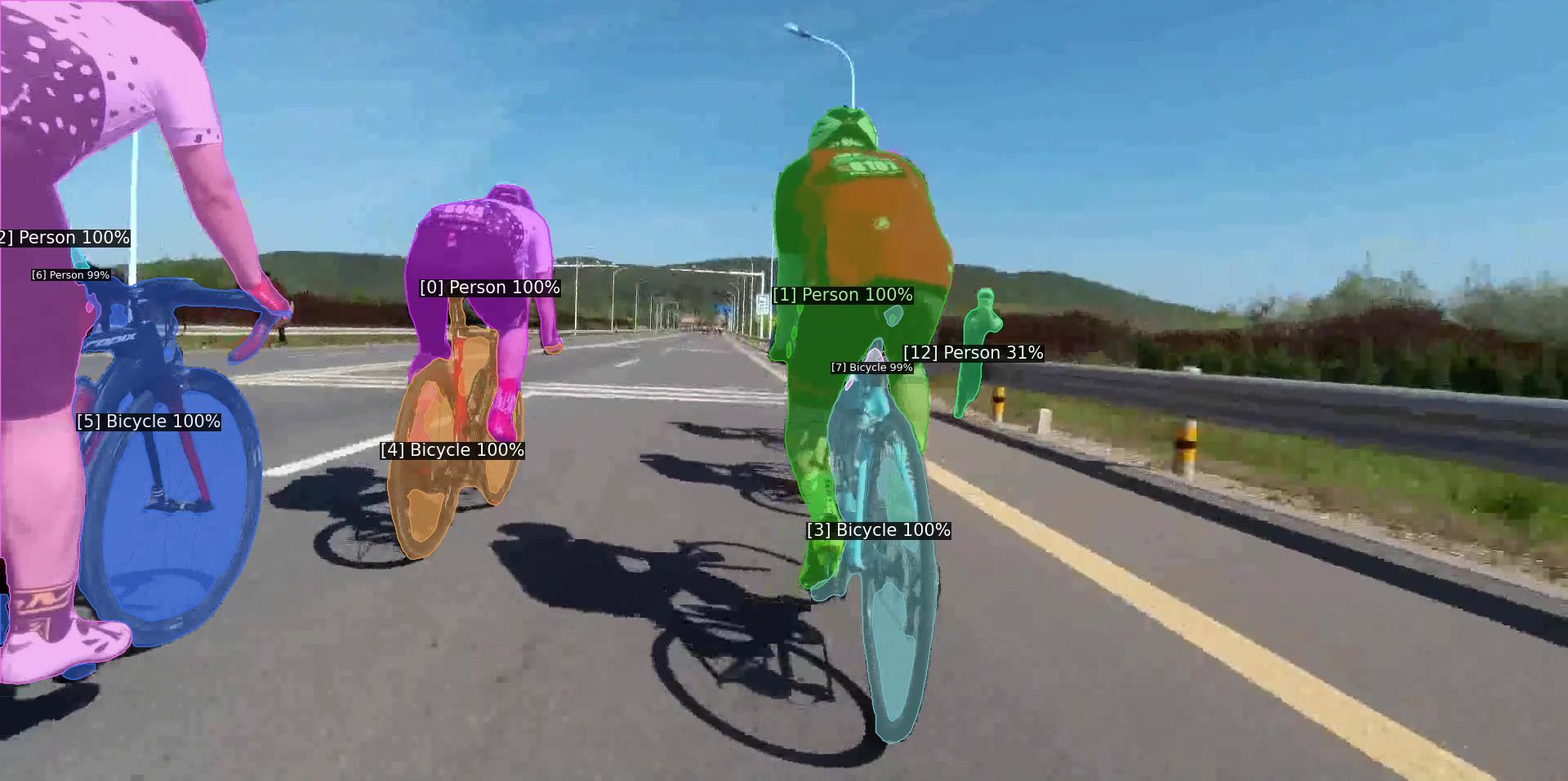}
    \includegraphics[width=\linewidth]{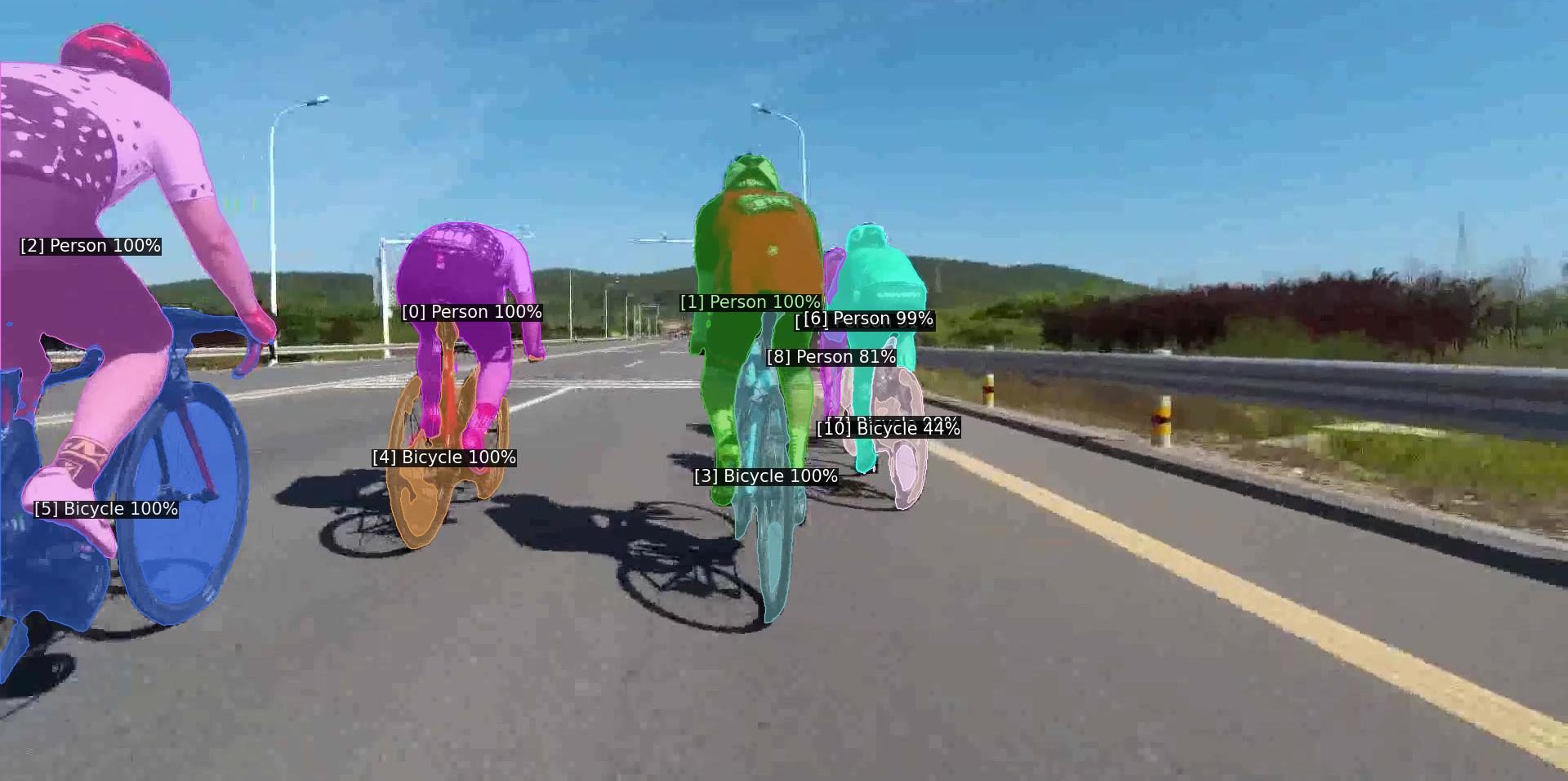}
    \includegraphics[width=\linewidth]{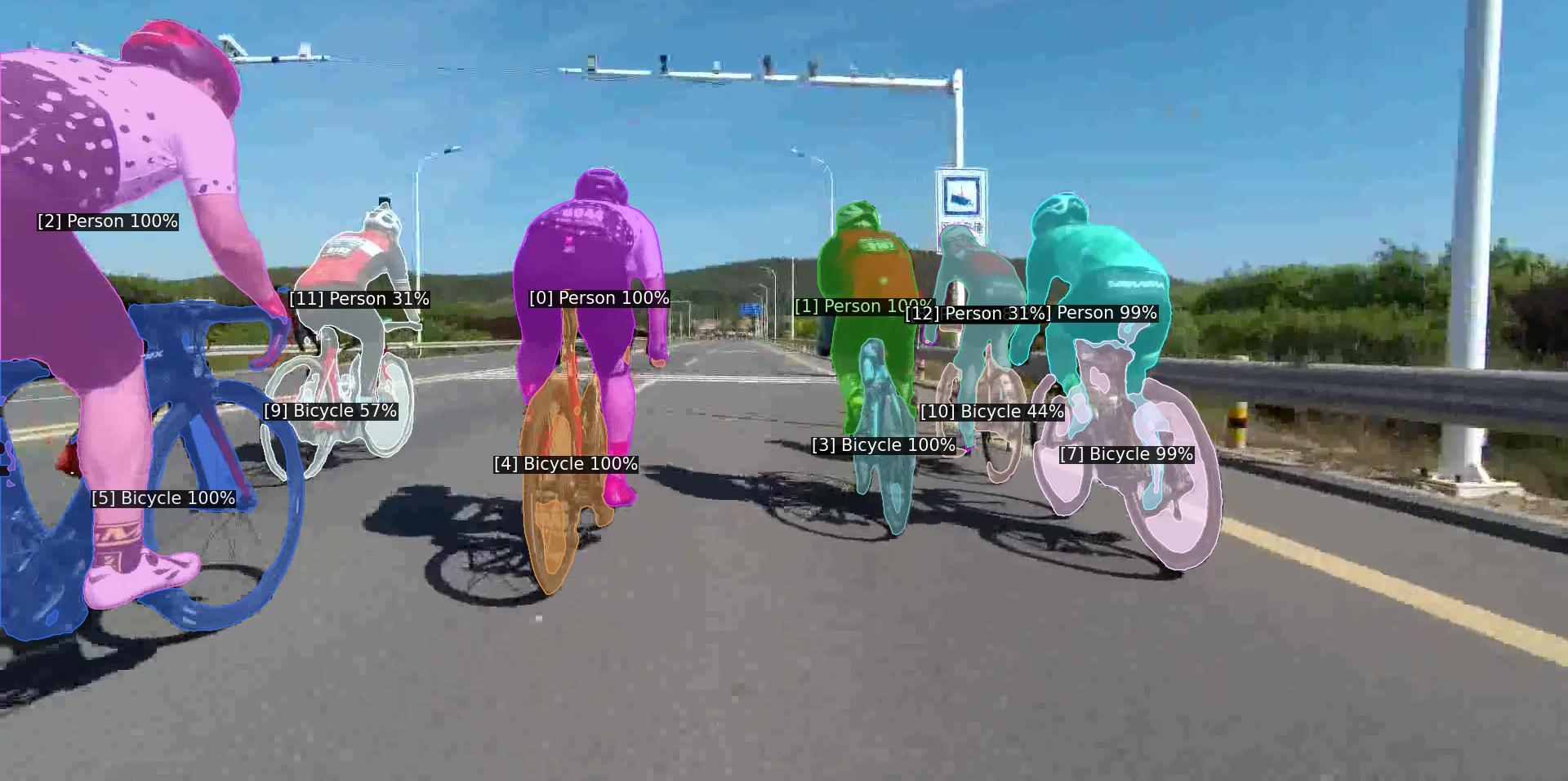}
    \includegraphics[width=\linewidth]{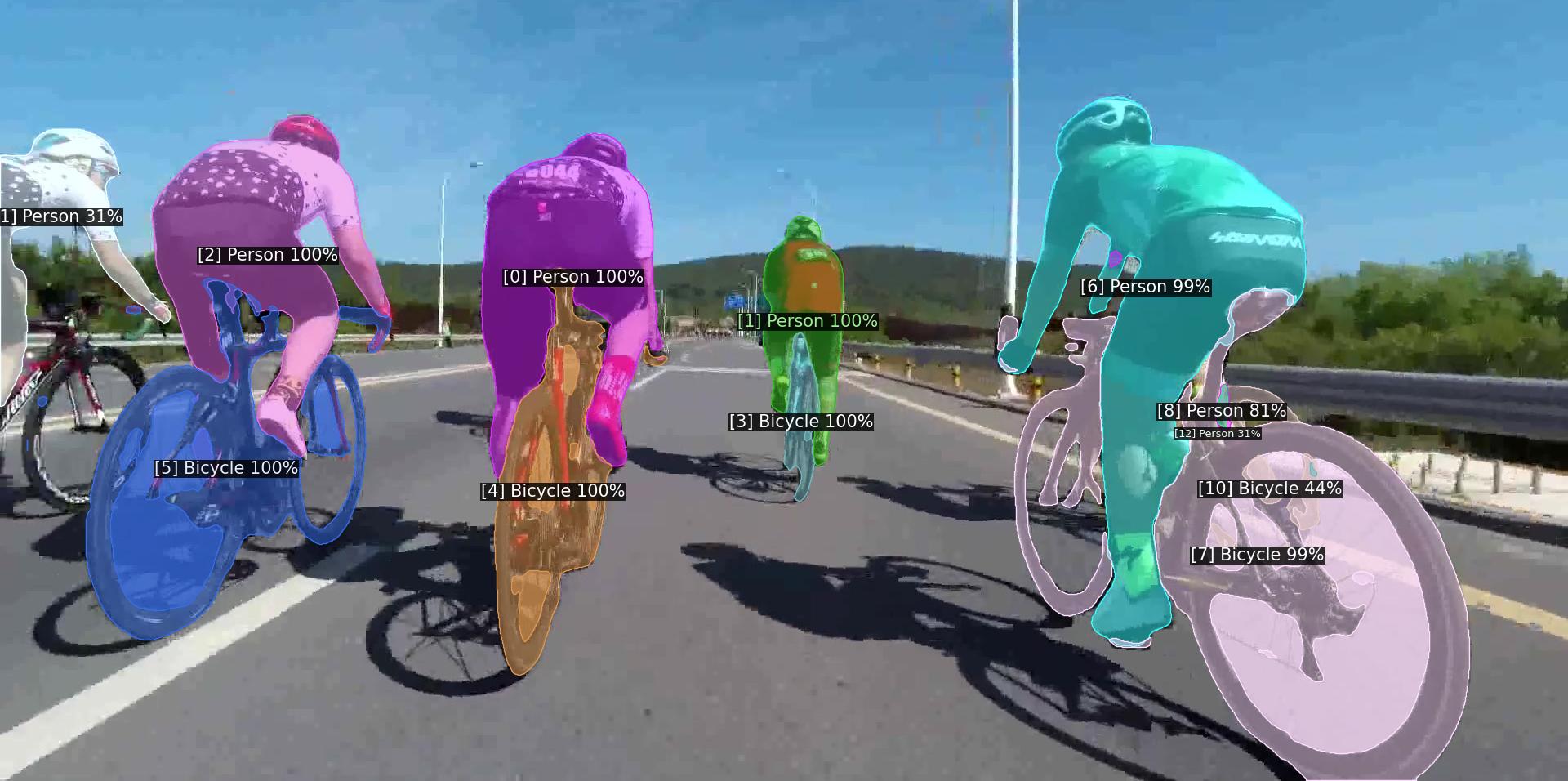}
    \includegraphics[width=\linewidth]{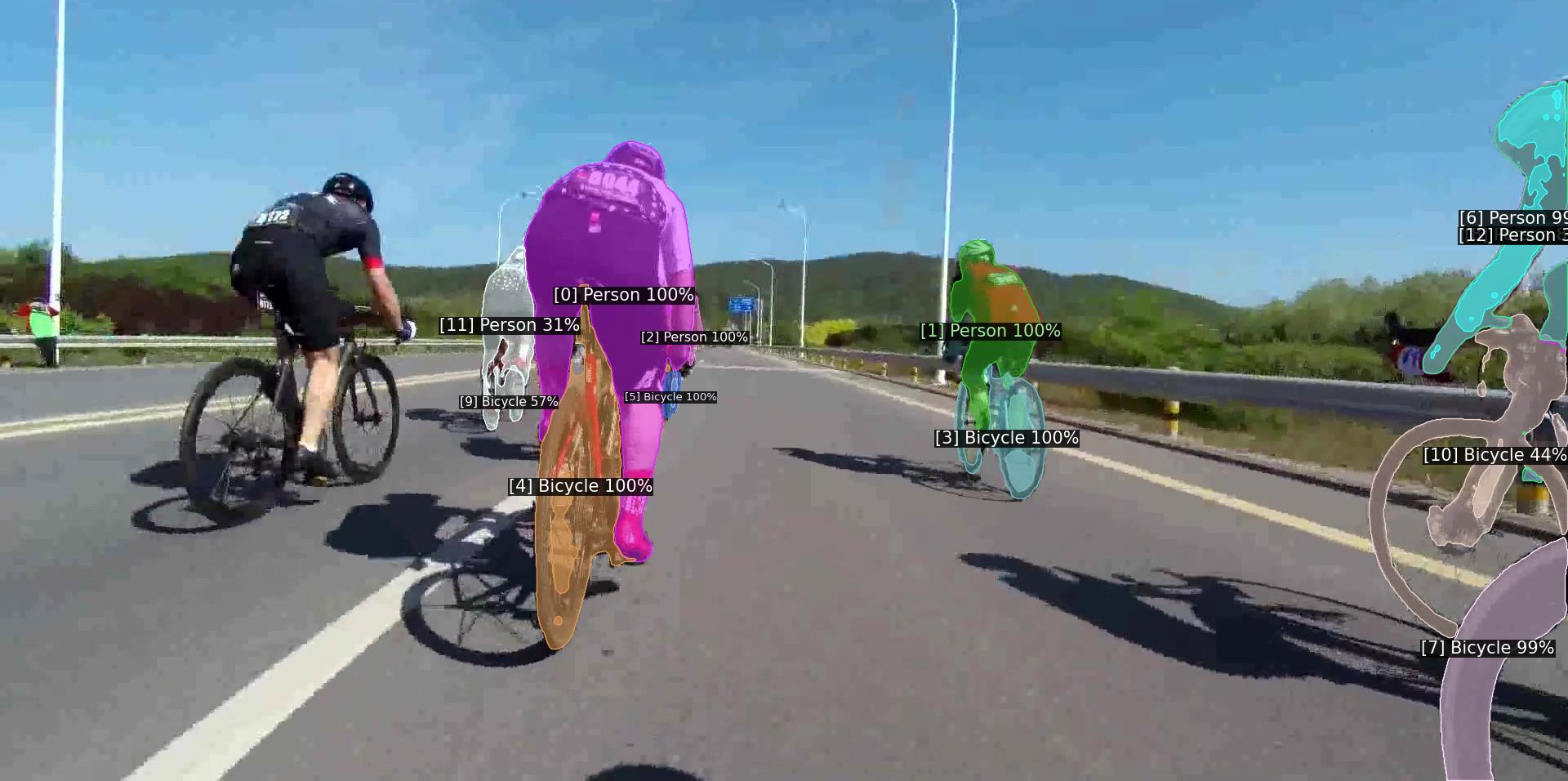}
    \caption*{\textbf{VidEoMT (112 FPS)}}
\end{minipage}
\caption{\textbf{Failure cases.} We illustrate a failure case of CAVIS~\citep{lee2025cavis} and VidEoMT on the OVIS dataset~\citep{qi2022occluded}. Both methods fail to detect newly appearing cyclists and bicycles in later frames.}
\label{app:fig:failure_case_2}
\end{figure*}

In \cref{app:fig:failure_case_1,app:fig:failure_case_2}, we present failure cases of CAVIS~\citep{lee2025cavis} and VidEoMT on the YouTube-VIS 2019~\citep{yang2019video} dataset. VidEoMT occasionally fails under long-term occlusions because query propagation is performed only between adjacent frames and because VidEoMT lacks a long-term memory mechanism (row 3 in \cref{app:fig:failure_case_1} and \cref{app:fig:failure_case_2}). It also struggles with newly appearing objects when the fixed set of queries has already been assigned to existing instances (row 6 in \cref{app:fig:failure_case_2}).

However, these limitations also apply to previous state-of-the-art methods, as they typically track objects only across consecutive frames and similarly lack long-term memory mechanisms. For instance, CAVIS also fails under long-term occlusions (row 4 in \cref{app:fig:failure_case_2}) and when new objects appear (row 3 in \cref{app:fig:failure_case_1} and rows 4–6 in \cref{app:fig:failure_case_2}).

\clearpage

{
    \small
    \bibliographystyle{ieeenat_fullname}
    \bibliography{main}
}